\documentclass[10pt,twocolumn,letterpaper]{article}

\usepackage[pagenumbers]{cvpr} 

\definecolor{cvprblue}{rgb}{0.21,0.49,0.74}
\usepackage[pagebackref,breaklinks,colorlinks,allcolors=cvprblue]{hyperref}
\usepackage[dvipsnames]{xcolor}     
\usepackage{colortbl} 
\usepackage{float}
\usepackage{verbatim} 
\usepackage{apalike}
\usepackage{amssymb}

\usepackage{colortbl} 
\usepackage{graphicx}
\usepackage{subcaption}

\usepackage{amsmath,bm}
\usepackage{comment}
\usepackage{makecell}
\usepackage{mathtools}
\usepackage{bbm}
\usepackage{enumitem}
\usepackage{multicol}
\usepackage{multirow}
\usepackage{csquotes}
\usepackage{wrapfig}
\usepackage{tablefootnote}
\usepackage{caption}

\usepackage{pifont}
\newcommand{\cmark}{\textcolor{PineGreen}{\ding{51}}}%
\newcommand{\xmark}{\textcolor{red}{\ding{55}}}%

\newcommand{\tab}{Table~}

\newcommand{\ourmodelsc}{DiffBlender}

\def\eqref#1{equation~\ref{#1}}

\def\1{\bm{1}}

\def\rmC{{\mathbf{C}}}

\def\rmI{{\mathbf{I}}}

\def\vc{{\bm{c}}}

\def\vh{{\bm{h}}}

\def\vx{{\bm{x}}}

\def\vz{{\bm{z}}}

\DeclareMathAlphabet{\mathsfit}{\encodingdefault}{\sfdefault}{m}{sl}
\SetMathAlphabet{\mathsfit}{bold}{\encodingdefault}{\sfdefault}{bx}{n}

\def\gL{{\mathcal{L}}}

\def\gN{{\mathcal{N}}}

\newcommand{\E}{\mathbb{E}}

\begin{document}

\title{ DiffBlender: Composable and Versatile\\Multimodal Text-to-Image Diffusion Models}

\author{
    Sungnyun Kim $^{1}$ $\quad$
    Junsoo Lee $^2$ $\quad$
    Kibeom Hong $^3$ $\quad$
    Daesik Kim $^2$ $\quad$
    Namhyuk Ahn $^4$  \\
    $^1$ KAIST AI \quad $^2$ NAVER WEBTOON AI \quad $^3$ Sookmyung Women's University \quad $^4$ Inha University \\
}

\twocolumn[{
\renewcommand\twocolumn[1][]{#1}
\maketitle
\begin{center}
\centering
\captionsetup{type=figure}
\vspace{-0.5em}
\centering
\vspace{-1em}
    \includegraphics[width=\linewidth]{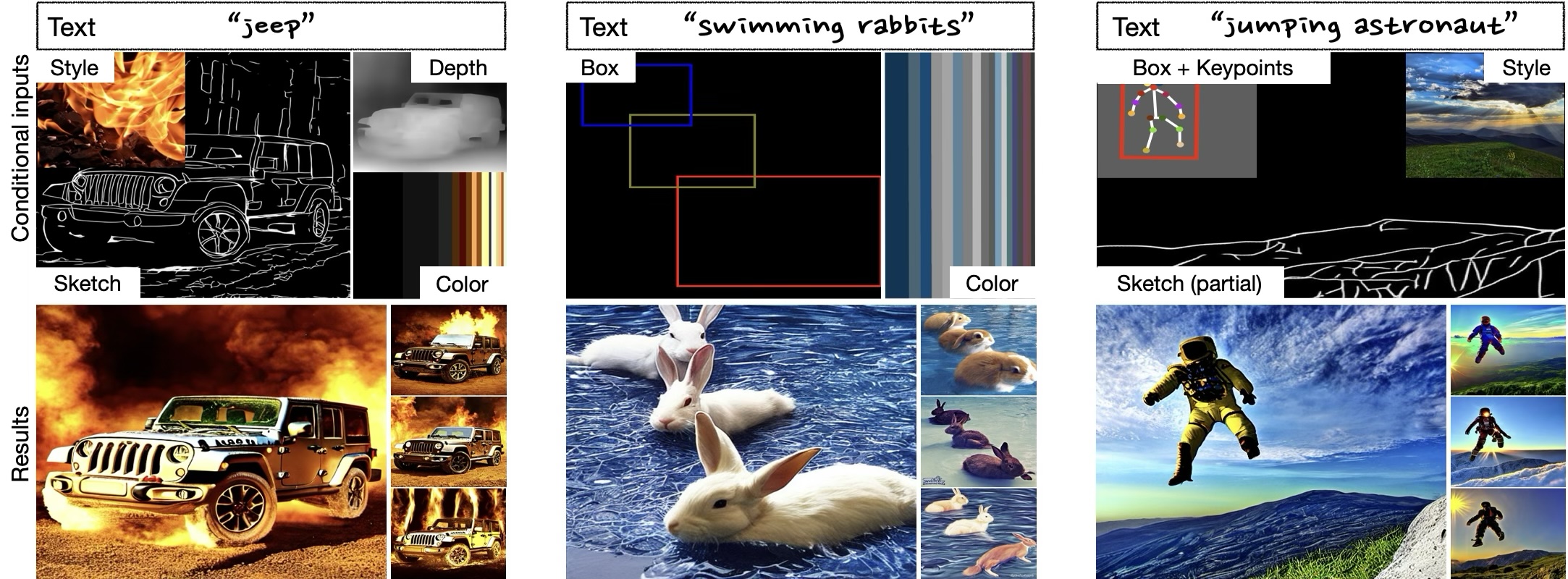}
    \vspace{-2em}
    \caption{\textbf{Generated images with multimodal conditions.} By incorporating various types of input modalities (1st row), {{\ourmodelsc}} successfully synthesizes high-fidelity and diverse samples, aligned with user preferences (2nd row).}
    \label{fig:first_teaser}
\label{fig:teaser}
\end{center}
}]

\begin{abstract}
In this study, we aim to enhance the capabilities of diffusion-based text-to-image (T2I) generation models by integrating diverse modalities beyond textual descriptions within a unified framework. To this end, we categorize widely used conditional inputs into three modality types: structure, layout, and attribute. We propose a multimodal T2I diffusion model, which is capable of processing all three modalities within a single architecture without modifying the parameters of the pre-trained diffusion model, as only a small subset of components is updated.
Our approach sets new benchmarks in multimodal generation through extensive quantitative and qualitative comparisons with existing conditional generation methods. We demonstrate that DiffBlender effectively integrates multiple sources of information and supports diverse applications in detailed image synthesis.
The code and demo are available at \url{https://github.com/sungnyun/diffblender}.
\end{abstract}

\section{Introduction}

In the field of image generation, significant progress has been driven by the strong generative capabilities of text-to-image (T2I) models based on diffusion processes~\cite{ho2020@ddpm, song2020ddim, nichol2021@improved_ddpm}. These advances are largely attributed to the availability of large-scale datasets that pair images with corresponding textual descriptions~\cite{schuhmann2022@laion, lin2014@mscoco, sharma2018@cc3m, ordonez2011@sbu}, as well as the development of aligned image-text embedding spaces~\cite{radford2021@clip}. Latent Diffusion Models (LDM)~\cite{rombach2022@ldm}, and their successor Stable Diffusion (SD), mitigate computational cost by operating in the latent space, while enabling flexible text conditioning via cross-attention on visual representations. Despite these advancements, relying solely on text prompts for fine-grained control remains a key limitation.

\begin{table*}[ht]
\centering
\caption{Summary of supported input modality types and training specifications for current methods---including joint training across multiple modalities and the extent of updated components. \textcolor{orange}{$\blacktriangle$} on \textit{partial training} indicates that their learnable parameter scale is greater than half but not the full SD network~\cite{rombach2022@ldm}.}
\vspace{-0.5em}
\label{tab:modality_comparison}
\setlength{\tabcolsep}{1.5em}
\begin{tabular}{l|ccc|c|c}
\hline
\multirow{2}{*}{Method} & \multicolumn{3}{c|}{Supported Modalities} & \multirow{2}{*}{\makecell{Multimodal \\ Training}} & \multirow{2}{*}{\makecell{Partial \\ Training}} \\
& Structure & Layout & Attribute & & \\
\hline\hline
GLIGEN~\cite{li2023@gligen}  & \xmark & \cmark & \xmark & \xmark & \cmark \\
ControlNet~\cite{zhang2023@controlnet} & \cmark & \xmark & \xmark & \xmark & \textcolor{YellowOrange}{$\blacktriangle$} \\
Uni-ControlNet~\cite{zhao2023unicontolnet} & \cmark & \xmark & \cmark & \cmark & \textcolor{YellowOrange}{$\blacktriangle$} \\
Composer~\cite{huang2023@composer} & \cmark & \xmark & \cmark & \cmark & \xmark \\
T2I-Adapter~\cite{mou2023@t2iadaptor} & \cmark & \xmark & \xmark & \xmark & \cmark \\
AnyControl~\cite{sun2025anycontrol} & \cmark & \xmark & \cmark & \cmark & \textcolor{YellowOrange}{$\blacktriangle$} \\
\hline
\ourmodelsc\ (Ours) & \cmark & \cmark & \cmark & \cmark & \cmark \\
\hline
\end{tabular}
\end{table*}

Incorporating additional modalities has been shown to significantly enhance image synthesis~\cite{zhang2023@controlnet}. However, existing methods encounter several limitations, as summarized in Table~\ref{tab:modality_comparison}. First, some models support only a single modality, which restricts their flexibility and limits their applicability across diverse scenarios~\cite{li2023@gligen, zhang2023@controlnet, mou2023@t2iadaptor}. Second, while some approaches enable multimodal conditioning, they often incur substantial computational costs due to the need to train separate adapters for each modality~\cite{mou2023@t2iadaptor, zhang2023@controlnet}. Third, certain methods rely on large-scale auxiliary modules, which significantly increase model complexity and resource requirements~\cite{huang2023@composer, zhao2023unicontolnet}.

To improve the usability and practicality of T2I models, these challenges must be addressed holistically. Specifically, future T2I systems should support diverse input modalities to better reflect user intent, enable efficient integration of multiple modalities through unified training, and minimize the size and overhead of additional modules required for multimodal conditioning.

Our objective is to develop a practical T2I model that can process multimodal inputs in a flexible and efficient manner. To this end, we categorize the most commonly used input conditions into three types: structure, layout, and attribute.
The structure modality represents the shape and composition of a scene, including inputs such as sketches, edge maps, and depth maps. The layout modality describes the spatial arrangement of objects using object-level coordinates, such as bounding boxes and keypoints. The attribute modality captures global characteristics of the scene, including color distributions and stylistic elements.
As shown in Table~\ref{tab:modality_comparison}, most existing T2I models support only one or two of these modality types. However, all three are essential, as they offer complementary information. For example, while the structure modality provides detailed and dense visual guidance, relying on it alone can impose a heavy burden on the user due to the need for precise input. Alternatively, layout information can convey spatial intent more succinctly, while structural data can be used to refine local details. This combination reduces user effort and improves controllability in achieving the desired output.

To this end, we propose a multimodal T2I synthesis model, referred to as \ourmodelsc, which accommodates diverse input modalities within a unified framework. Unlike prior approaches, \ourmodelsc\ is both composable and versatile—it supports all three modality types within a single architecture. This design allows users to flexibly combine input conditions according to their intent, thereby providing enhanced controllability over the generation process.
Table~\ref{tab:modality_comparison} presents a comparison of existing diffusion-based T2I models. \ourmodelsc\ supports all modality types and enables efficient multimodal training. Notably, while some recent methods~\cite{zhao2023unicontolnet,huang2023@composer} adopt multimodal conditioning, they do not support layout inputs and require extensive training resources, which limits their usability. In contrast, \ourmodelsc\ uniquely supports all three modality types in an efficient and unified manner.
Figure~\ref{fig:first_teaser} illustrates generated examples from \ourmodelsc\ under various combinations of input conditions, demonstrating its ability to synthesize high-quality images that align with user-specified preferences.

In addition, \ourmodelsc\ achieves high computational efficiency. By adopting a unified framework and training strategy, it updates only a small subset of components while keeping the entire set of Stable Diffusion (SD) parameters frozen. This significantly reduces the number of additional parameters required for handling multimodal conditions and lowers the dependency on large-scale training data. In contrast to existing models that rely on massive datasets, our approach demonstrates strong performance even when trained on relatively modest datasets (e.g., COCO~\cite{lin2014@mscoco}).

To fulfil these objectives, we incorporate a \textit{Blender block} into the downsampling blocks of the UNet architecture in Stable Diffusion (SD). This addition enables seamless processing of multimodal conditions within a unified framework. The Blender block employs a shared design across modalities, avoiding modality-specific structures. While the original self-attention and cross-attention modules in SD’s UNet remain frozen, only the lightweight Blender components are updated during training.
To address potential conflicts between modalities during inference, we introduce a mode-specific guidance mechanism. This technique enables precise and independent control over each modality, facilitating smooth and balanced integration. Our main contributions are summarized as follows.

\begin{itemize}[leftmargin=*, label={$\circ$}]
\item We propose {\ourmodelsc}, a novel diffusion model that effectively handles complex combinations of modalities, carefully curated and categorized into three distinct types.
\item We design the model to extend seamlessly to additional modalities without requiring expensive training for the  flexible usage.
\item We introduce a novel guidance that provides precise control over specific modalities, a critical feature for multimodal generation.
\item Quantitative and qualitative evaluations, along with a human perception test, demonstrate that {\ourmodelsc} achieves high fidelity and reliable generation under multiple conditions.
\end{itemize}

\section{Related Work}
\label{sec:related_work}

\textbf{Diffusion models.}
These have emerged as a powerful approach for generating high-quality images~\cite{ho2020@ddpm, song2020ddim, nichol2021@improved_ddpm, nichol2021@glide, saharia2022image, saharia2022palette, ruiz2022@dreambooth, chang2023muse, tang2024any, choi2023finecontrolnet}. They outperform adversarial models~\cite{esser2021taming, karras2019@stylegan, sauer2021@projectedgan, xu2018@attngan, ye2021improving, tao2022@dfgan} in both fidelity and diversity~\cite{dhariwal2021@adm}. Among these, Latent Diffusion Models (LDM)~\cite{rombach2022@ldm} provide a unified and effective approach for conditional generation, which later evolved into Stable Diffusion (SD).
Subsequent studies have advanced T2I synthesis by incorporating large language models~\cite{saharia2022photorealistic@imagan, li2022upainting, zhang2023controllable, chen2025textdiffuser}, leveraging CLIP’s image-text joint representation space~\cite{radford2021@clip, ramesh2022hierarchical}, or using a Mixture-of-Experts structure~\cite{xue2024raphael, feng2023ernie}. In this work, we use SD as the backbone diffusion model and extend its capabilities to support a wide range of applications.

\noindent\textbf{Multimodal conditioning.}
Expressing fine-grained details (\eg spatial layout) through text prompts alone is challenging. To address this limitation, several studies have extended T2I models to receive additional inputs, such as sketches~\cite{voynov2022sketch-diffusion}, depth maps~\cite{huang2023@composer, zhang2023@controlnet}, bounding boxes~\cite{zhao2019image, sun2019image, zhou2024migc, wang2024instancediffusion}, reference images~\cite{yuan2022text, kwon2022diffusion}, color palettes~\cite{huang2023@composer}, scene graphs~\cite{johnson2018image, yang2022diffusion}, semantic maps~\cite{couairon2022diffedit, wang2022pretraining, li2021collaging, park2019@spade}, or combinations of concepts~\cite{liu2022compositional, wang2023compositional}.

Among these, Composer~\cite{huang2023@composer} supports diverse modalities within a single model but requires training the entire model from scratch, which results in high training costs. ControlNet~\cite{zhang2023@controlnet} and Uni-ControlNet~\cite{zhao2023unicontolnet} leverage the generative prior of SD but require large amounts of data to train a significant number of additional parameters. This approach limits scalability for adding new modalities. GLIGEN~\cite{li2023@gligen} and T2I-Adapter~\cite{mou2023@t2iadaptor} focus on partial updates of conditioning modules to achieve more efficient conditional image generation. However, they can support only a single modality type, mainly image-based conditions.
In this work, we contribute by designing SD-based models that support a broader range of conditions. To this end, we categorize the key conditioning modalities and provide a unified framework that efficiently handles all of them.

\begin{figure}[t]
    \centering
    \includegraphics[width=.9\linewidth]{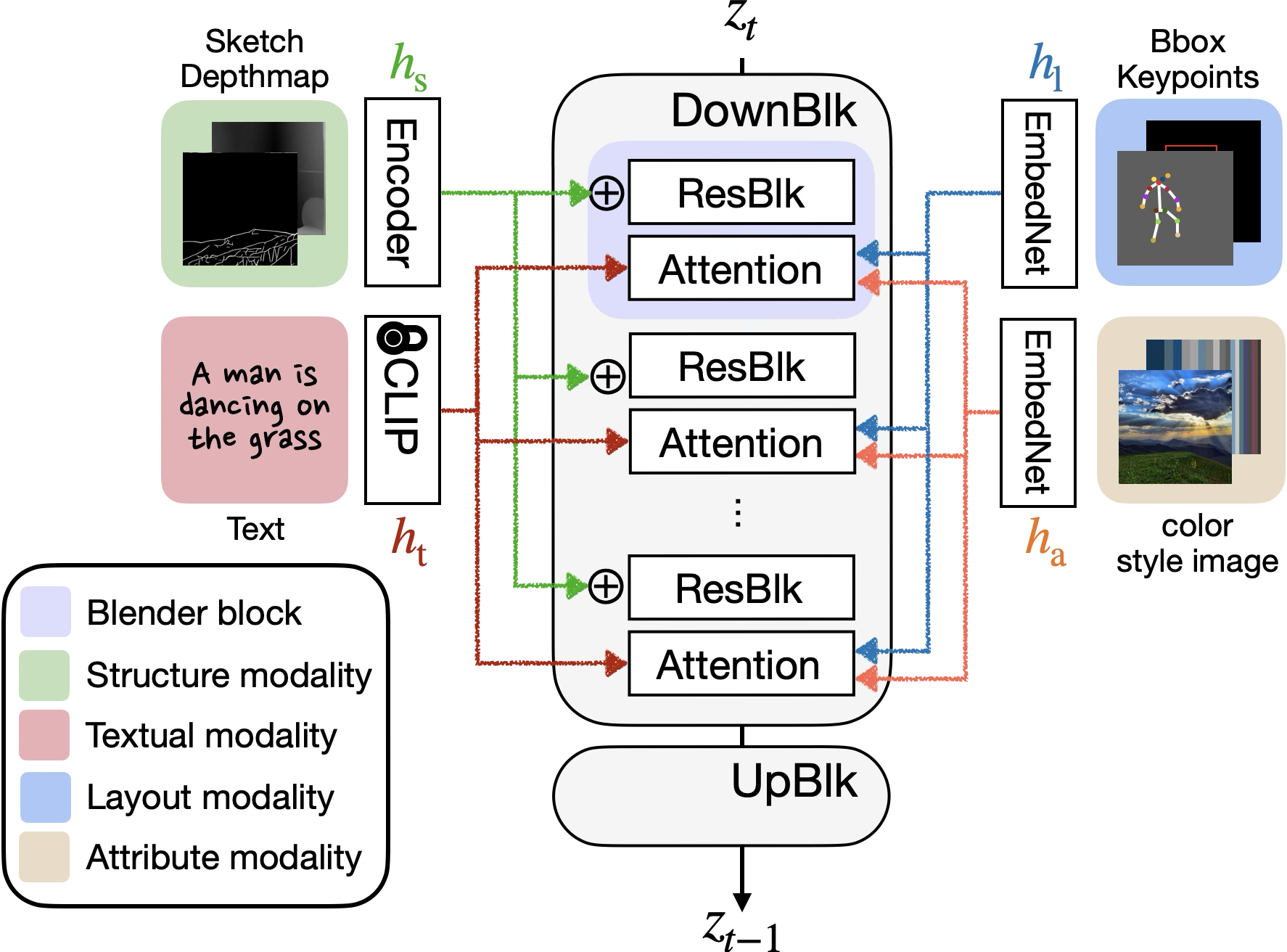}
    \caption{
    \textbf{Model Overview.} \ourmodelsc\ accepts all modality types—text, structure, layout, and attribute—through the Blender block. This design allows to operate within a unified framework using only a small number of additional parameters. Note that layout and attribute, although visually represented as images, are processed in tokenized (or vectorized) forms.}
    \label{fig:model_architecture}
    \vspace{-1.2em}
\end{figure}

\section{Method}

\noindent\textbf{Model Overview.}
DiffBlender is a unified multimodal text-to-image generation model built upon Stable Diffusion~\cite{rombach2022@ldm}. It supports three major input modality types—structure, layout, and attribute, which are processed through corresponding blending modules (Section~\ref{sec:modality}). These are integrated into the UNet backbone via lightweight additions (Section~\ref{subsec:method_model_architecture}), without modifying the original pretrained parameters. As illustrated in Figure~\ref{fig:model_architecture}, the model accepts multimodal inputs and injects them into the diffusion process via modulated self-attention within the Blender blocks (Section~\ref{sec:blender}). This design enables compositional control while maintaining high generation quality with minimal computational overhead.

\subsection{Preliminary: Latent Diffusion Models (LDM)}
\label{subsec:method_preliminaries}

In LDM~\cite{rombach2022@ldm}, a bi-directional mapping network is trained to produce the latent representation $\vz$ of the image $\vx$, \ie $\vz=\text{enc}(\vx), \tilde{\vx}=\text{dec}(\vz)$. Then, a UNet-based~\cite{ronneberger2015unet} denoiser is trained as a diffusion model on the latent $\vz$.
Starting from the noise-induced $\vz_T$ and conditioning text $\vc$, the diffusion model $f_{\bm \theta}$ generates less noisy samples, gradually producing $\vz_{T-1}$ to $\vz_0$.
The training objective of LDM is thus described as follows:
\begin{equation}\label{eq:ldm}
    \min_{\bm\theta} \gL_{\text{LDM}} = \E_{\vz,{\bm \epsilon} \sim \gN(\mathbf{0},\rmI), t} \big[ \| {\bm \epsilon} - f_{\bm\theta} (\vz_t, t, \vc)\|_2^2 \big],
\end{equation}
where $t$ is uniformly sampled from time steps $\{1,\cdots,T\}$, $\vz_t$ is the step-$t$ noisy of $\vz$, and $f_{\bm\theta}$ is the diffusion model. 
The denoiser is built using ResNet~\cite{he2016deep@resnet} and Transformer~\cite{vaswani2017attention@transformer} blocks. It predicts the noise $\hat{\bm \epsilon}$, which matches the size of the input $\vz$. The time embedding is injected into each ResNet block, and the text $\vc$ is conditioned to cross-attention layers.

\subsection{Supported Modality}
\label{sec:modality}
To ensure both usability and flexibility, \ourmodelsc\ is designed to accommodate user intent while supporting a broad spectrum of input modalities. Seamless integration of multimodal inputs requires a structured organization of diverse conditional signals. To this end, we categorize the inputs into three primary modality types—Structure, Layout, and Attribute—which together encompass the most prevalent forms of input used in conditional image synthesis tasks.

$\bullet$~The \textit{structure} modality provides detailed, image-based representations of a scene, serving as a strong source of guidance during synthesis. It includes inputs such as sketches and depth maps, which convey precise spatial and compositional information. While structure-based inputs can substantially improve image quality, they are often difficult for users to produce, which limits their practicality when used in isolation.

$\bullet$~The \textit{layout} modality mitigates the usability challenges of the structure modality by providing simplified spatial cues, such as bounding boxes. These inputs are easier to define and manipulate, making them more accessible to users. However, due to their sparsity, layout inputs alone may result in outputs that lack the visual fidelity achieved with structure-based guidance.

$\bullet$~The \textit{attribute} modality captures global properties of a scene, including color palettes and stylistic characteristics. Unlike structure and layout, attribute inputs do not specify spatial arrangements but instead influence the overall tone and appearance of the generated image, allowing for high-level aesthetic control.

These three modality types are deliberately defined to balance their distinct strengths and limitations. Most existing conditional T2I models support only a single modality, such as structure~\cite{mou2023@t2iadaptor} or layout~\cite{li2023@gligen}, which limits their overall usability. In contrast, \ourmodelsc\ enables users to flexibly combine and control multiple modalities, providing a greater degree of customization. For instance, users can selectively apply structural guidance to specific regions or combine layout and attribute conditions to achieve the desired output.
This composable and versatile design sets \ourmodelsc\ apart from conventional T2I models, establishing it as a more powerful and user-friendly framework for a wide range of application scenarios.

\subsection{Model Architecture}
\label{subsec:method_model_architecture}
As illustrated in Figure~\ref{fig:model_architecture}, \ourmodelsc\ processes the defined modality types within a unified architecture. Built upon Stable Diffusion (SD), it replaces the inner components of the downsampling blocks in the UNet with Blender blocks, which integrate the input conditions into the latent feature space of SD.
To improve training efficiency, we avoid duplicating UNet structures, as seen in prior work~\cite{zhang2023@controlnet,sun2025anycontrol}. Although such replication can provide performance benefits, it poses challenges when supporting diverse modalities and demands substantial training data. Instead, \ourmodelsc\ employs modality-specific designs informed by the predefined modality categories, enabling effective support for a wide range of input types.
To streamline the incorporation of layout and attribute conditions, we represent these modalities in tokenized form. In contrast, structure inputs, which are image-based, are encoded into latent representations and directly injected into the Blender block. Layout and attribute inputs are first embedded through dedicated networks before being injected into the model.

\begin{figure}[!t]
    \centering
\includegraphics[width=0.8\linewidth]{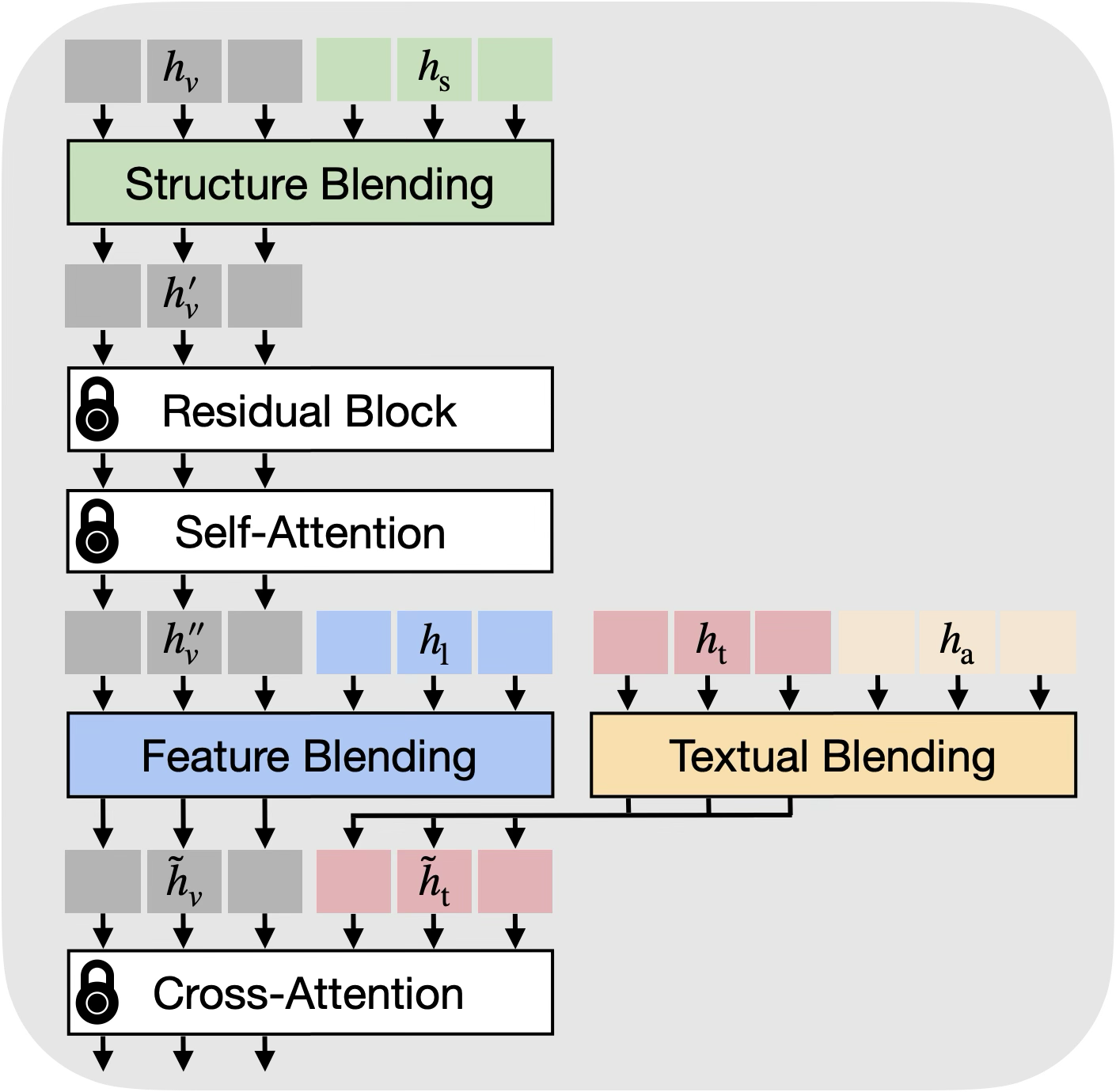}
    \caption{\textbf{Blender block.}
Leveraging the residual and attention blocks of Stable Diffusion, we inject and blend multimodal conditions through a series of specialized blending modules. The Structure Blending Module (SBM) maps structure information onto the latent visual features, which are subsequently processed by the residual and self-attention layers. The Feature Blending Module (FBM) integrates layout information into the visual tokens, facilitating more accurate spatial alignment within the synthesized scene. The Textual Blending Module (TBM) projects global attribute inputs onto the textual embedding space, enabling high-level control over style and appearance. Layers indicated with a lock symbol denote components that remain frozen.
}
\vspace{-1em}
\label{fig:blender_block}
\end{figure}

\subsection{Blender Block}
\label{sec:blender}
As illustrated in Figure~\ref{fig:blender_block}, our Blender block is built upon SD’s basic building block. In the original SD network, each block receives visual tokens $\vh_v$ and text embedding tokens $\vh_t$, which are processed through residual blocks, self-attention, and cross-attention layers before being passed to the next block.
We freeze these original blocks and add three blending modules, each designed to handle a specific modality type. Through these blending modules, conditional information is projected onto $\vh_v$ and $\vh_t$.

The \textit{structure blending module (SBM)} incorporates structure modality conditions. Structure inputs are encoded into latent representations $\vh_{s,k}$ for the $k$-th image-form structure input using a ResNet-based image encoder. We blend the visual latent $\vh_v$ with structure latent $\vh_{s}$ as: $\vh_v' = \vh_v + \sum_k \vh_{s,k}$.
This blending process is applied at four downsampling blocks of the diffusion UNet, enabling effective incorporation of structure information into the latent space.
The updated latent $\vh_v^{\prime}$, now containing the projected structure information, is passed through the original residual block and self-attention layer, producing the refined latent feature $\vh_v''$.

With a given $i$-th layout modality condition $\vh_{l,i}$, we modulate $\vh_v''$ through the \textit{feature blending module (FBM)}. Unlike structure, which uses image-form conditions, layout inputs are not provided in image form. Inspired by \cite{li2023@gligen}, we instead adopt coordinate tokens as the input format to enhance usability and editability. The user-provided layout conditions are converted into layout tokens through an embedding network.
These tokens, which contain spatial information, directly influence the latent feature $\vh_v''$ and can modify the position or orientation of objects. This enables users to easily assign object placements or define poses in the generated samples.

To project the layout modality $\vh_{l,i}$ into the latent feature $\vh_v''$, we concatenate and blend them through a self-attention layer:
\begin{equation}
\tilde{\vh}_v'' = \text{SA} \big( \big[ \vh_v''\,; \{\vh_{l,i}\} \big] \big),
\end{equation}
where $[;]$ denotes concatenation.
Although the fusion begins with a simple concatenation of $\vh_v''$ and the layout tokens $\{\vh_{l,i}\}$, the subsequent self-attention layer is fundamentally responsible for the actual integration of information across modalities. Specifically, given the combined sequence $X = \big[ \vh_v''\,; \{\vh_{l,i}\} \big]$, self-attention computes pairwise interactions between all token pairs using learned attention weights, effectively modeling complex inter-dependencies:
\begin{equation}
    \text{SA}(X) = \text{softmax}\left( \frac{QK^\top}{\sqrt{d_k}} \right)V,
\end{equation}
where $Q = XW_Q,\; K = XW_K,\; V = XW_V$. Each token, including both layout and visual tokens, attends to every other token in the sequence. This mechanism allows the model to reason jointly over spatial layout and visual context. As a result, the self-attention operation inherently performs content-aware fusion based on the relationships among all tokens.

We then modulate the original latent feature with the self-attended blending feature as follows:
\begin{equation}
\tilde{\vh}_v =\vh_v'' + \text{tanh}(\beta_l) \cdot \tilde{\vh}_v''[:V],
\label{eq:sab}
\end{equation}
where $V$ is the length of the visual tokens. This formulation gradually imposes conditions on the intact visual tokens, gated by a learnable parameter $\beta_l$.
Through the FBM, the latent feature and layout modality information are effectively aligned.
Note that we retain only the first $V$ tokens corresponding to the visual latent positions. This design is inspired by recent practices in attention-based architectures~\cite{alayrac2022@flamingo,li2023@gligen}. The goal is to apply modulation selectively to spatial regions of interest, in line with the residual connection structure where the self-attended features must match the original feature dimensions. This strategy helps maintain stability during both training and inference.

The \textit{Textual blending module (TBM)} injects the attribute modality by modulating it into the textual embedding instead of the visual token. This approach avoids conflicts with the conditioning prompt, as the attribute modality primarily contains global and abstract information. Without this design, either the text prompt or the attribute condition could be disregarded. Additionally, directly projecting the attribute modality onto the visual features makes it difficult to naturally reflect the overall appearance and tone. By projecting the attribute modality onto the textual embedding, the original $\vh_t$ is effectively adapted to the attribute condition input.
For simplicity and efficiency, we design the TBM to function similarly to the FBM:
\begin{equation}
    \tilde{\vh}_{t} = \vh_t + \text{tanh}(\beta_t) \cdot \text{SA} \Bigl( \big[ \vh_{t}\,; \{\vh_{a,j}\} \big] \Bigr)[:L],
\end{equation}
where $\vh_{a,j}$ are tokens for the $j$-th attribute condition, $\beta_t$ is a learnable gate paramter, and $L$ is the textual token length. Consequently, the visual features are delivered to the next layer after cross-attending two intermediate features: $\vh_v = \text{CA}(\tilde{\vh}_v, \tilde{\vh}_t)$.

\subsection{Mode-Specific Guidance for Decoupled Control}
\label{subsec:method_mode_specific_guidance}
Classifier-free guidance (CFG)~\cite{ho2022@cfg} is a simple yet effective method for improving conditional image generation~\cite{nichol2021@glide, saharia2022photorealistic@imagan}. It adjusts the predicted noise by including the gradient of the log-likelihood of $p(\rmC|\vx_t)$.
However, for multimodal control, CFG applies the same level of guidance to all conditions, which prevents it from addressing scenarios where users need to control only a subset of modalities.
To address this limitation, we extend CFG by proposing mode-specific guidance (MSG), which controls the influence of specific modalities and provides precise control over certain conditional inputs.

Formally, given that $p(\rmC|\vx_t) \propto p(\vx_t|\rmC)/p(\vx_t)$, we can obtain $\nabla_{\vx_t} \!\log p(\rmC|\vx_t) \propto {\bm \epsilon}^*(\vx_t, \rmC) - {\bm \epsilon}^*(\vx_t)$, when we know their exact scores. Instead, $\nabla_{\vx_t} p(\vx_t|\rmC)$ and $\nabla_{\vx_t} p(\vx_t)$ are parameterized via score estimators $\hat{\bm \epsilon}(\vx_t, \rmC)$ and $\hat{\bm \epsilon}(\vx_t, \emptyset)$, which correspond to the conditional and unconditional noise estimators, respectively. These are derived from the denoising diffusion model. The adjusted noise prediction is expressed as $
\hat{\bm \epsilon}^*(\vx_t, \rmC) = \hat{\bm \epsilon}(\vx_t, \emptyset) + w (\hat{\bm \epsilon}(\vx_t, \rmC) - \hat{\bm \epsilon}(\vx_t, \emptyset))$,
where $w$ is the guidance scale.

To control the influence of a specific modality $\vc_m$, we need to control the estimate of $\nabla_{\vx_t} \log p(\vc_m | \vx_t, \rmC \setminus \vc_m)$, which is given as
\[
p(\vc_m | \vx_t, \rmC \setminus \vc_m) = \frac{p(\vc_m, \vx_t | \rmC \setminus \vc_m)}{p(\vx_t | \rmC \setminus \vc_m)} \propto \frac{p(\vx_t | \rmC)}{p(\vx_t | \rmC \setminus \vc_m)}.
\]
Thus, the MSG direction becomes $\hat{\bm \epsilon}(\vx_t, \rmC) - \hat{\bm \epsilon}(\vx_t, \rmC \setminus \vc_m)$, where the second term is derived by using null input for $\vc_m$, given the other modalities. The adjusted noise prediction for $\vc_m$, combined with the original CFG, is expressed as
\begin{equation*}
    \hat{\bm \epsilon}_m^*(\vx_t, \rmC) = \hat{\bm \epsilon}^*(\vx_t, \rmC) + \gamma \bigl(\hat{\bm \epsilon}(\vx_t, \rmC) - \hat{\bm \epsilon}(\vx_t, \rmC \setminus \vc_m)\bigl),
\end{equation*}
where $\gamma \in \mathbb{R}$ is the scale factor for MSG, which controls the impact of $\vc_m$, and $\gamma=0$ reverts to the original CFG.

\section{Experimental Setups}

\subsection{Training}
The objective function is similar to that of LDM, but we replace $\vc$ in Eq.~\ref{eq:ldm} with $\rmC = \{ \vc_\text{text}, \vc_\text{sketch}, \vc_\text{box}, \cdots \}$ to incorporate a set of modalities. The model learns to denoise the latent $\vz_t$ at a given time step $t$ using all conditions in $\rmC$.
While naive joint training of multiple modalities is effective, careful scheduling based on modality properties can further improve performance. For example, the structure modality provides detailed scene structure, potentially making the (less detailed) layout modality redundant. This redundancy can hinder the model’s ability to understand all modalities. To address this, we train \ourmodelsc\ in a sparse-to-dense approach. Initially, we train with sparse information (\eg layout, attributes) and introduce dense information (\eg structure) in the later stages. This strategy enhances controllability while improving the model’s understanding of different modalities.

\subsection{Implementation Details}
\label{supp_sec:implementation_details}

We train \ourmodelsc\ on the widely used COCO2017~\cite{lin2014@mscoco} training set, which contains approximately 120,000 images. In contrast to prior T2I methods that depend on web-scale datasets such as LAION~\cite{schuhmann2022@laion}, \ourmodelsc\ demonstrates effective performance using a substantially smaller dataset.
Training is conducted with a batch size of 32 for 200,000 iterations. We employ a learning rate of 5e-5 with the AdamW optimizer~\cite{loshchilov2017@adamw}, incorporating a 10,000-step warm-up phase. Input images are center-cropped to a resolution of 512×512. For classifier-free guidance (CFG)~\cite{ho2022@cfg}, we randomly drop \textit{structure} conditions with a probability of 50\%, and all other modalities with a probability of 10

\begin{table}[!t]
\centering
\caption{\textbf{Analysis of training strategy}. We analyze the training process by varying the introduction of conditional modalities during the early (75\% of training) and the late stages (last 25\%).}
\vspace{-0.5em}
\label{tab:training_ablation}

\setlength{\tabcolsep}{0.5em}
\begin{tabular}{l|cc|cc}
\hline
Strategy & Early stage & Late stage & Layout $\uparrow$ & FID $\downarrow$  \\
\hline\hline
Config A & S+L+A & S+L+A & 13.1 & 14.8 \\
Config B & L+A & S+L+A & \textbf{15.8} & 15.0 \\
Config C & L+A & S & 15.6 & \textbf{14.1} \\
\hline
\end{tabular}
\vspace{-1em}
\end{table}

We construct the multimodal conditioning inputs as follows:
	1) \textit{Structure} modality: This includes sketches and depth maps. Sketches are extracted using PiDiNet~\cite{su2021@pidinet} and subsequently simplified through a sketch refinement process~\cite{simo2018@sketchsimplification}. Depth maps are generated via monocular depth estimation~\cite{ranftl2020@depthmap}.
	2)	\textit{Layout} modality: We utilize annotated grounding boxes and human keypoints from the COCO2017 dataset.
	3)	\textit{Attribute} modality: This includes color and reference style or content. To extract color statistics, we employ a smoothed CIELab histogram~\cite{sergey@rayleigh}, with a palette space defined by 11 hue bins, 5 saturation levels, and 5 lightness levels, using a smoothing parameter of $\sigma=10$. Reference style or content is captured by encoding input images into feature embeddings using the CLIP image encoder~\cite{radford2021@clip}.

\subsubsection{Embedding Networks}
For the structure modality, we design dedicated encoder networks to process inputs such as sketches and depth maps. Each encoder consists of eight ResNet blocks, with each block comprising two convolutional layers. Downsampling operations are applied at the 3rd, 5th, and 7th blocks to maintain compatibility with the visual feature resolution of the original UNet. At the 2nd, 4th, 6th, and 8th blocks, latent features are extracted through a zero-conv module~\cite{zhang2023@controlnet}, enabling progressive influence over the latent fusion process.
Prior to being processed by these encoders, sketches and depth maps are first encoded using the autoencoder-\textit{KL}~\cite{rombach2022@ldm}, which shares the same pretrained encoder as the input image. Importantly, these autoencoders remain frozen throughout training.

For the layout modality, we adopt the COCO2017~\cite{lin2014@mscoco} annotation format and follow the implementation provided by GLIGEN~\cite{li2023@gligen}. Layout conditions consist of grounding boxes and human keypoints, with each image containing up to 30 bounding boxes and 8×17 keypoints (17 per person, for up to 8 individuals).
Each bounding box is associated with a textual label, which is embedded using the CLIP text encoder~\cite{radford2021@clip}, along with a 4-dimensional coordinate vector encoded via Fourier embedding. The resulting embeddings are concatenated and passed through a three-layer MLP with a hidden dimension of 512 and an output dimension of 768. Similarly, each keypoint includes a person identifier and 2D coordinates, which are embedded using person-specific embeddings and Fourier embeddings. These are also concatenated and processed through a three-layer MLP with identical dimensions.

\begin{figure*}[t]
    \centering
    \includegraphics[width=\linewidth]{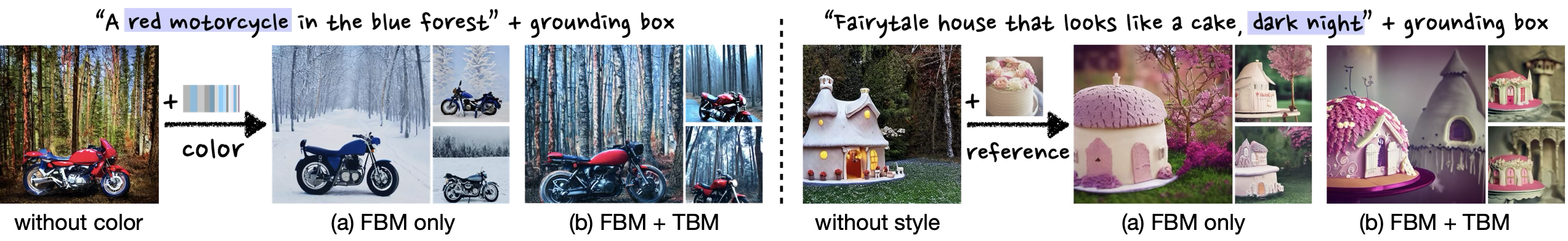}
    \vspace{-1.5em}
    \caption{
    \textbf{Analysis on the Blending modules.} (a) Injecting the attribute modality through FBM (feature blending module) fails to express the delicate color and style of the reference. (b) Introducing TBM (textual blending module) effectively reflects the attribute modality.
    }
    
    \label{fig:lsa_gsa}
\end{figure*}

\begin{figure*}[t]
    \centering
    \includegraphics[width=\linewidth]{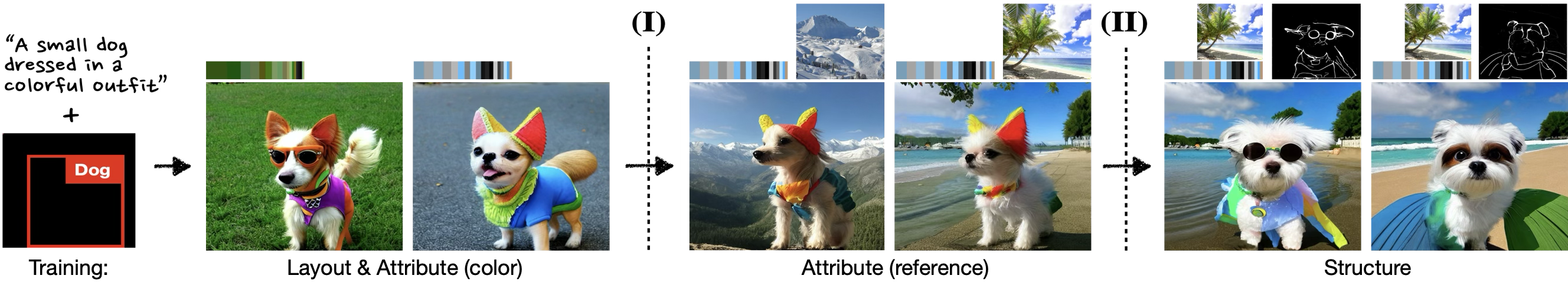}
    \vspace{-1.5em}
    \caption{
    \textbf{Continual modality training.} We simulate a scenario where users introduce new modalities to \ourmodelsc. With its composable module design, \ourmodelsc\ handles this efficiently without requiring re-training of the entire model, unlike previous methods.
    }
    
    \label{fig:modality_extension}
    \vspace{-1em}
\end{figure*}

For the attribute modality, we design dedicated embedding networks to handle both color palettes and reference images. Both networks employ a three-layer MLP with LayerNorm and SiLU activation functions, featuring a hidden dimension of 512 and an output dimension of 768. Reference images are first encoded using the CLIP image encoder, and the resulting embeddings are then passed through the MLP. For color palettes, histogram values are scaled by a factor of 10 to address their low magnitude, and subsequently encoded using Fourier embedding. This scaling helps accelerate convergence during training.

\begin{table}[t]
\caption{Performance analysis when varying the input conditions.}
\label{tab:additional_quantitative}
\vspace{-0.5em}
\centering
\setlength{\tabcolsep}{0.3em}
\begin{tabular}{l|cccc}
\hline
Input Condition & Layout $\uparrow$ & Structure $\uparrow$ & Depth $\uparrow$ & FID\,$\downarrow$ \\
\hline\hline
Box & 15.8 & - & - & 19.6 \\
Box + Color & 15.6 & - & - & 19.6 \\
Box + Reference & 14.8 & - & - & 17.5 \\
\hline
Sketch & - & 33.5 & 87.9 & 16.7 \\
Depth & - & 30.6 & 88.1 & 16.4 \\
Sketch + Depth & - & 33.0 & 88.3 & 16.9 \\
\hline
All & 19.7 & 34.8 & 88.5 & \textbf{14.1} \\
All + MSG (Ours) & \textbf{20.1} & \textbf{37.6} & \textbf{88.8} & 17.0 \\
\hline
\end{tabular}
\vspace{-1em}
\end{table}

\subsubsection{Hyperparameters}
During training, we apply standard data augmentations to the input images, including random horizontal flipping and center cropping to a resolution of 512×512. To enhance robustness and enable conditioning on partial structural inputs (e.g., partial sketches), segmentation maps are used to randomly mask regions of the sketch. The probability of applying segmentation-based masking is set to 0.7, with an equal likelihood (0.5) of masking either the foreground or background.
For classifier-free guidance (CFG)~\cite{ho2022@cfg}, we apply a drop ratio of 0.5 for the structure modality and 0.1 for all other modalities. Captions are also randomly dropped with a probability of 0.1, and null inputs are used to replace the omitted conditions. Within Transformer blocks, learnable null tokens initialized to zero are employed for dropped textual or tokenized inputs, while non-trainable zero-filled tensors are used for image-based conditions.

During inference, two hyperparameters come into play: classifier-free guidance scale\,($w$) and scheduling coefficient $\alpha$. We use $w=5.0$ by default. $\alpha$ serves as scaling coefficients to schedule the conditioning strength during the inference. $\alpha_{s}$ is multiplied after the structure blending module, $\alpha_{l}$ is multiplied after the feature blending module, and $\alpha_{a}$ is multiplied after the textual blending module, respectively.
When $T$ represents the total number of denoising diffusion steps, the values of $\alpha_{l}$ and $\alpha_{a}$ are set to 1 for the first $0.3\,T$ steps, and then set to 0 for the remaining $0.7\,T$ steps. $\alpha_{s}$ remains at a constant value of 0.7 for all $T$ steps. These hyperparameters were determined based on achieving the lowest FID score.

\subsubsection{Mode-Specific Guidance}
The CFG scale~\cite{ho2022@cfg}, $w$, is conventionally set as $[5.0, 20.0]$, empirically found in the literature~\cite{rombach2022@ldm}. Because our MSG is added to the original CFG, we set $\gamma$ no less than $-w$ value; otherwise, it will not only reduce the impact of modality but also degrade the image quality. Conversely, if we set $\gamma$ too high, the image generation will highly depend on the specified modality. In our experiments, we use $w=5.0$ and $\gamma \in [-3.0, 3.0]$.

\begin{figure*}[t]
    \centering
    \includegraphics[width=1.0\linewidth]{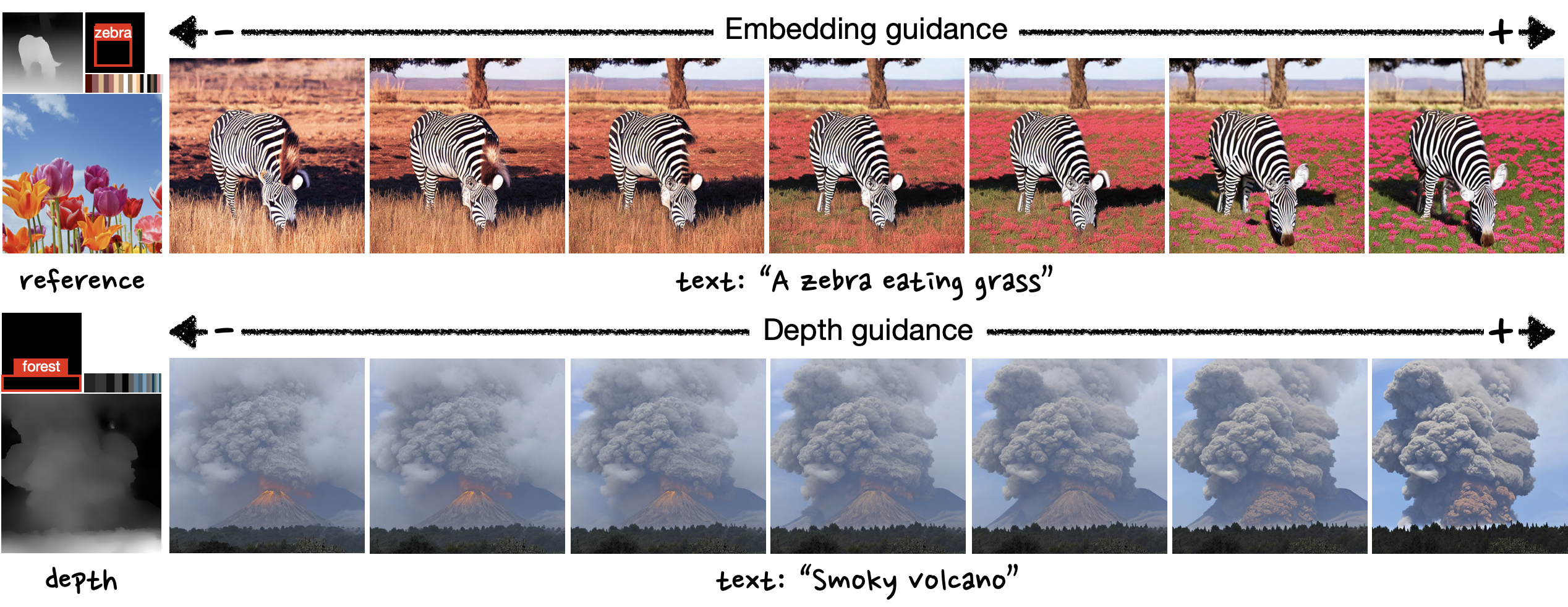}
    \vspace{-1.5em}
    \caption{
    \textbf{Mode-specific guidance.}
    We linearly change the MSG scale $\gamma$, where the center indicates $\gamma\!=\!0$\,(original CFG).
    }
    \vspace{-1em}
    \label{fig:modality_guidance}
\end{figure*}

\begin{figure*}[!h]
    \centering
    \includegraphics[width=\linewidth]{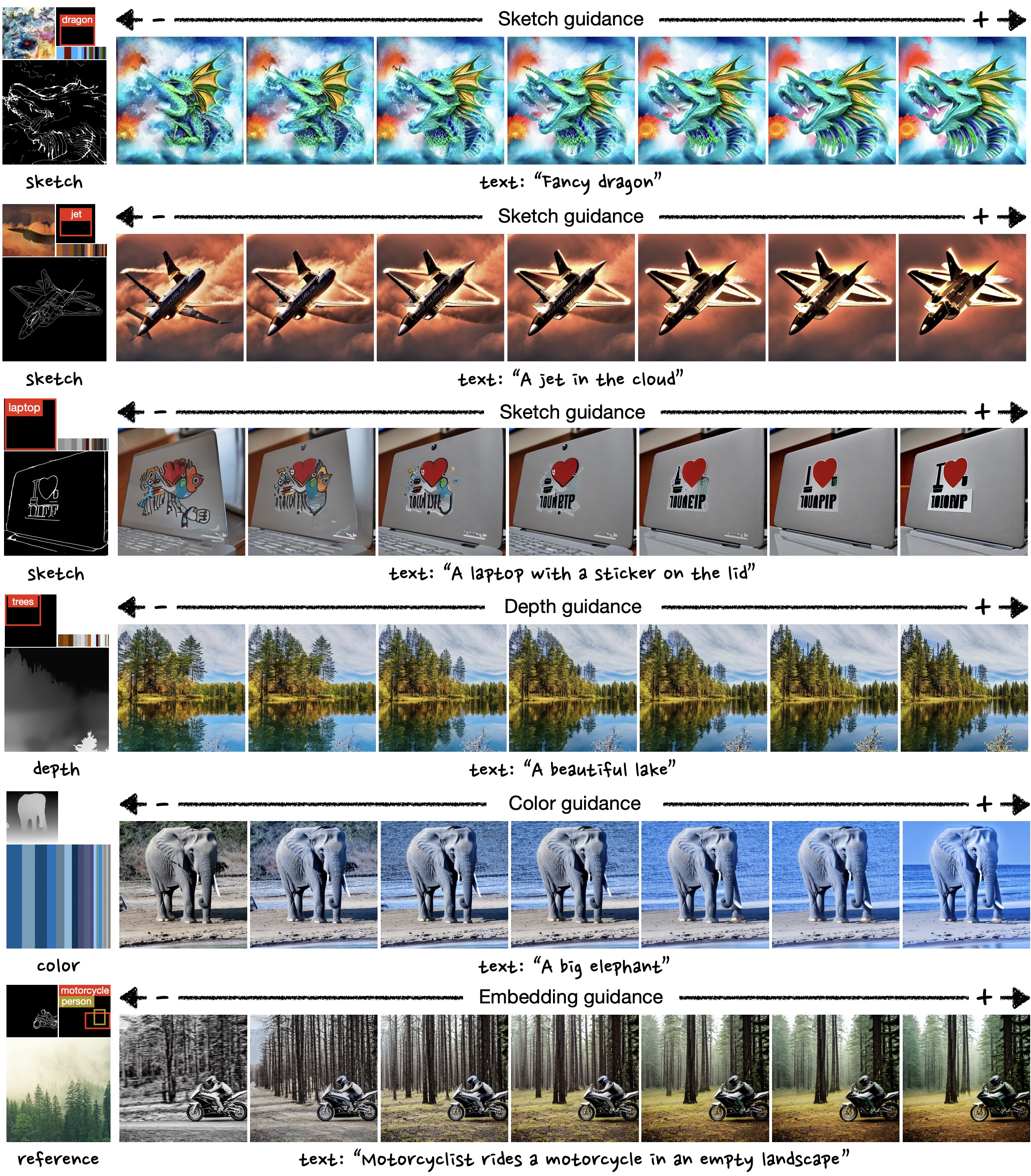}
    \vspace{-1.5em}
    \caption{\textbf{Mode-specific guidance.} Each row depicts the mode-specific guided samples for each modality. We linearly change the scale $\gamma$, where the center indicates $\gamma=0$ (original classifier-free guidance).
    }
    \label{fig:appx_modality_guidance}
    \vspace{-1em}
\end{figure*}

\section{Results and Discussions}
In this section, we present the main evaluation results of DiffBlender. We compare our model with existing conditional T2I baselines across multiple metrics, analyze its controllability under different modality conditions, and highlight practical applications of our method.

\subsection{Evaluation Metrics}
We evaluate our model using 5,000 samples from the COCO2017~\cite{lin2014@mscoco} validation set, cropped to a resolution of 512$^2$ pixels. Our evaluation focuses on both the fidelity and controllability of the generated images from multiple perspectives.
For fidelity, we employ standard metrics commonly used in T2I synthesis tasks. These include the Fréchet Inception Distance (FID)\cite{heusel2017@fid} to assess overall image quality, and the CLIP score\cite{radford2021@clip} to measure alignment with textual descriptions. However, these metrics alone are insufficient to fully evaluate performance in multimodal generation settings.

To address this limitation, we introduce additional metrics tailored to specific input modalities. The \textit{Layout score} evaluates the spatial correspondence between the generated image and the input layout condition, computed using YOLO-v5~\cite{jocher2022@yolov5} with $\text{AP}_{50}$. The \textit{Structure score} measures structural alignment by calculating SSIM~\cite{wang2004@ssim} between the sketches of generated and ground-truth images. Likewise, the \textit{Depth score} quantifies depth consistency using the formula $100 - (10 \times \ell_2\text{-distance with depth maps})$.

\subsection{Model Analysis}
\label{subsec:analysis}

\noindent\textbf{Blending module.}
In the design of \ourmodelsc, the Textual Blending Module (TBM) modulates the textual embeddings based on attribute modality inputs. By decoupling the roles of structure and layout modalities from that of the attribute modality, \ourmodelsc\ is able to reflect attribute-driven information more accurately.
As shown in Figure~\ref{fig:lsa_gsa}, we train a baseline model that injects all modalities through the Feature Blending Module (FBM) alone to demonstrate the necessity of a dedicated module for attribute processing. This FBM-only configuration often overemphasizes color or reference embeddings at the expense of textual prompts (e.g., overriding prompt details such as the motorcycle’s color or the phrase “dark night”). In contrast, incorporating TBM allows for more precise control by modulating global features at the textual level, resulting in outputs that better adhere to the given prompts.
This analysis underscores the importance of recognizing the distinct roles of each modality and leveraging appropriate attention mechanisms to achieve well-conditioned image synthesis.

\begin{table}[t]
\centering
\caption{Comparison of token selection strategies in the feature blending module.}
\vspace{-0.5em}
\setlength{\tabcolsep}{1em}
\begin{tabular}{l|cc}
\hline
\textbf{Token Selection Strategy} & Layout $\uparrow$ & Depth $\uparrow$ \\
\hline\hline
All tokens $\rightarrow$ interpolation & 18.4 & 86.4 \\
Random token sampling               & 18.0 & 86.1 \\
\hline
Interest region tokens (Ours)       & \textbf{19.7} & \textbf{88.5} \\

\hline
\end{tabular}
\label{tab:token_selection}
\vspace{-1em}
\end{table}

\begin{table}[t]
\centering
\caption{Ablation study on the structure blending module (SBM).}
\vspace{-0.5em}
\setlength{\tabcolsep}{0.3em}
\begin{tabular}{l|ccc}
\hline
\textbf{Model Variant} & Layout $\uparrow$ & Structure $\uparrow$ & Depth $\uparrow$ \\
\hline\hline
DiffBlender w/o SBM            & 15.8 & 31.4 & 87.5 \\
DiffBlender w/ SBM (full)        & \textbf{19.7} & \textbf{34.8} & \textbf{88.5} \\
\hline
\end{tabular}
\label{tab:sbm}
\vspace{-1em}
\end{table}

\noindent\textbf{Continual modality training.}
An ideal multimodal T2I model should be capable of seamlessly adapting to new modalities without requiring modifications to the underlying architecture. However, existing approaches often face difficulties in such scenarios due to the large number of parameters that must be retrained and the architectural inflexibility in accommodating additional modalities.

In contrast, \ourmodelsc\ is designed with a flexible and extensible architecture that efficiently supports all commonly used modality types—Structure, Layout, and Attribute—as defined in our framework. As illustrated in Figure~\ref{fig:modality_extension}, \ourmodelsc\ enables continual modality integration with minimal computational overhead, while preserving previously learned conditional behaviors. This composable design allows users to incorporate new modalities on demand by updating only partial components, without retraining the entire model. As a result, the model can be easily extended to support customized inputs, further improving its adaptability and effectiveness.

\begin{figure*}[t]
    \centering
    \begin{subfigure}{.325\textwidth}
        \centering
        \includegraphics[width=\linewidth]{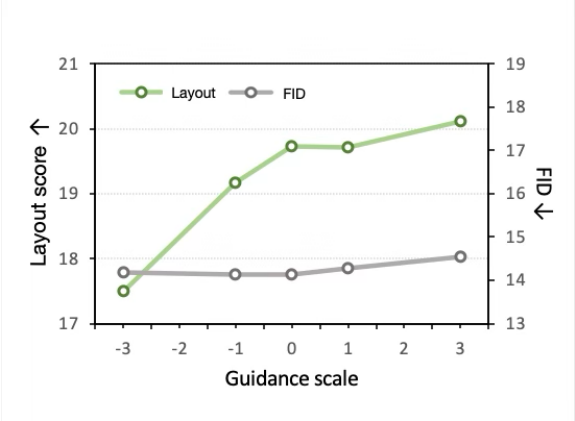}   
        \caption{Grounding box guidance}\label{fig:msg_box}
    \end{subfigure}
    \begin{subfigure}{.325\textwidth}
        \centering
        \includegraphics[width=\linewidth]{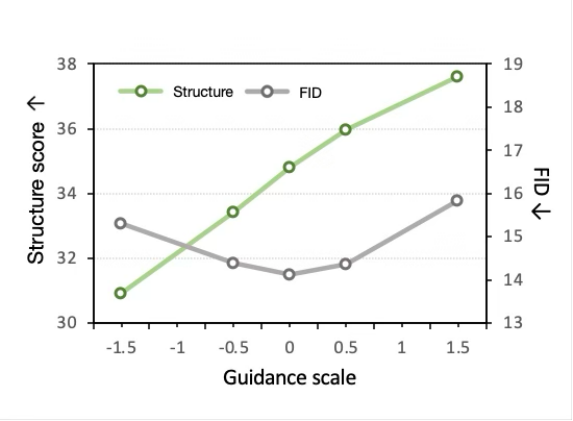}
        \caption{Sketch guidance}\label{fig:msg_sketch}
    \end{subfigure}
    \begin{subfigure}{.325\textwidth}
        \centering
        \includegraphics[width=\linewidth,]{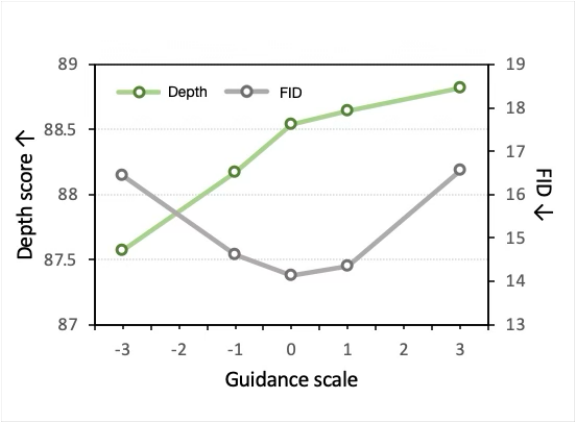}  
        \caption{Depth map guidance}\label{fig:msg_depth}
    \end{subfigure}
    \vspace{-0.5em}
    \caption{\textbf{Analysis on mode-specific guidance control, measured on COCO2017 validation set.}}
    \label{fig:msg}
\end{figure*}

\begin{figure*}[ht]
    \centering
    \includegraphics[width=\linewidth]{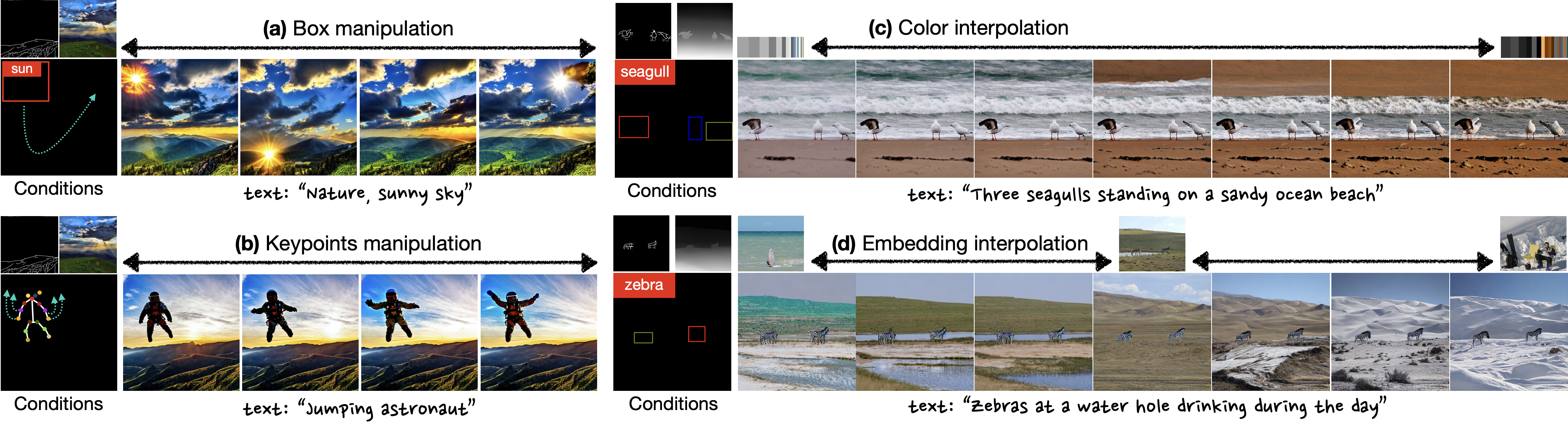}
    
    \caption{
    \textbf{Modality manipulation.} In this scenario, we edit the layout and attribute modality conditions. The intermediate resulting images consistently reflect the changes and transition smoothly.
    }

    \label{fig:latent_interpolation}
    \vspace{-1em}
\end{figure*}

\begin{figure*}[!h]
    \centering
    \includegraphics[width=\linewidth]{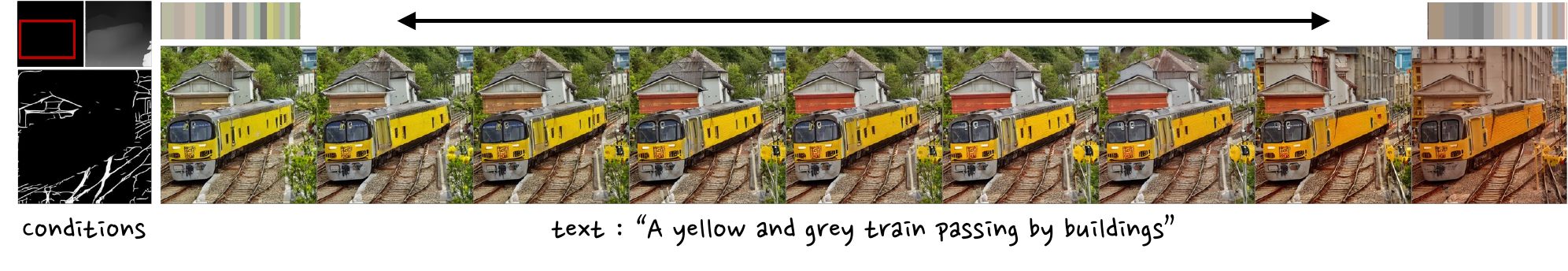}
    \includegraphics[width=\linewidth]{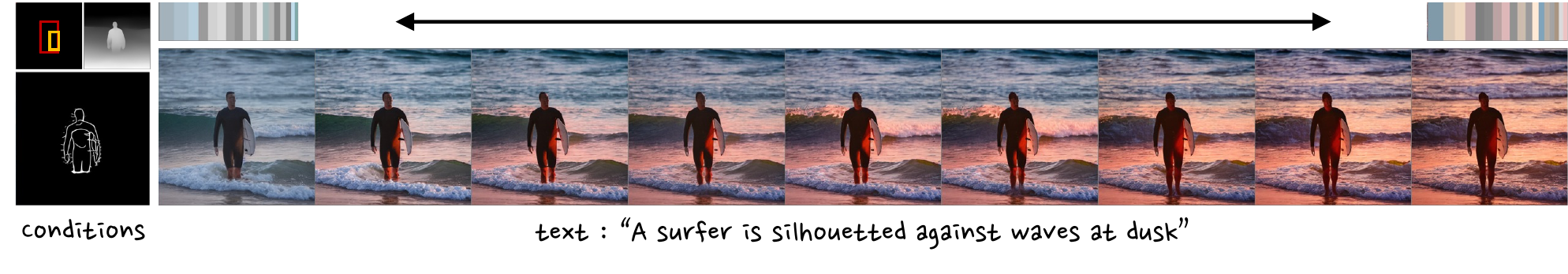}
    \includegraphics[width=\linewidth]{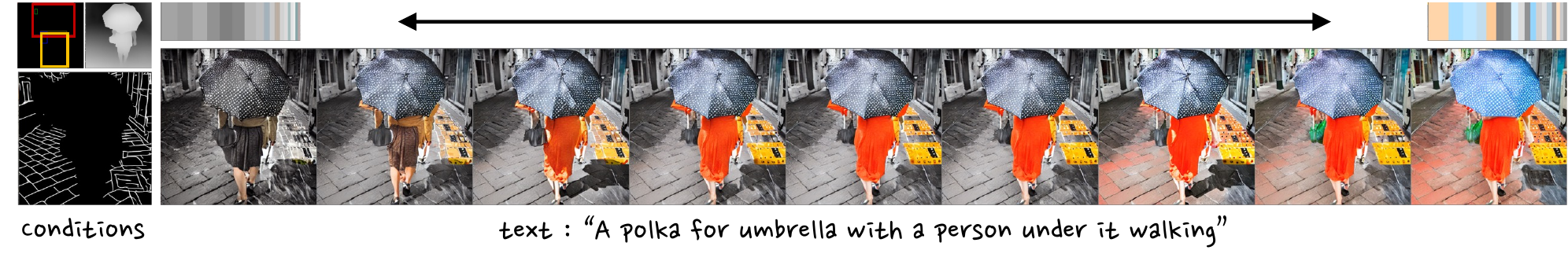}
    \includegraphics[width=\linewidth]{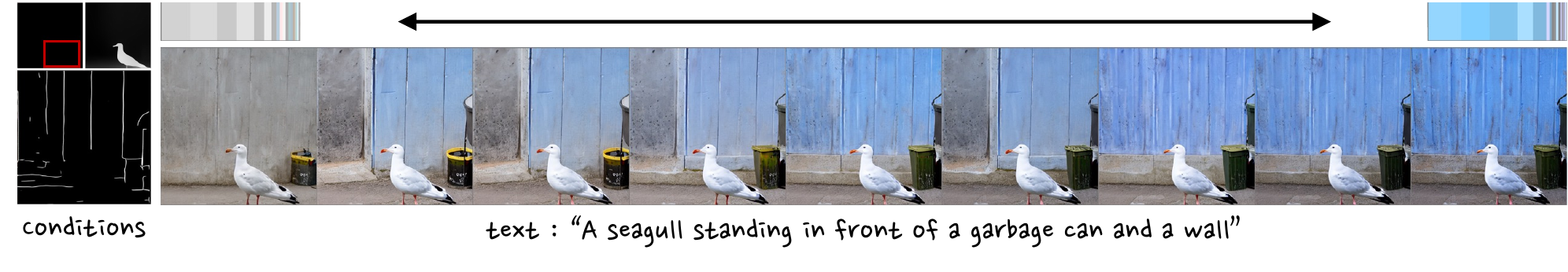}
    \includegraphics[width=\linewidth]{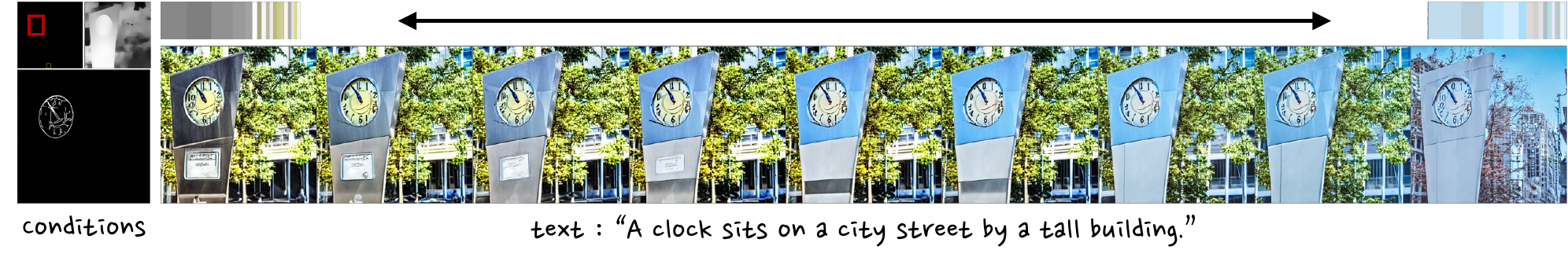}
    \includegraphics[width=\linewidth]{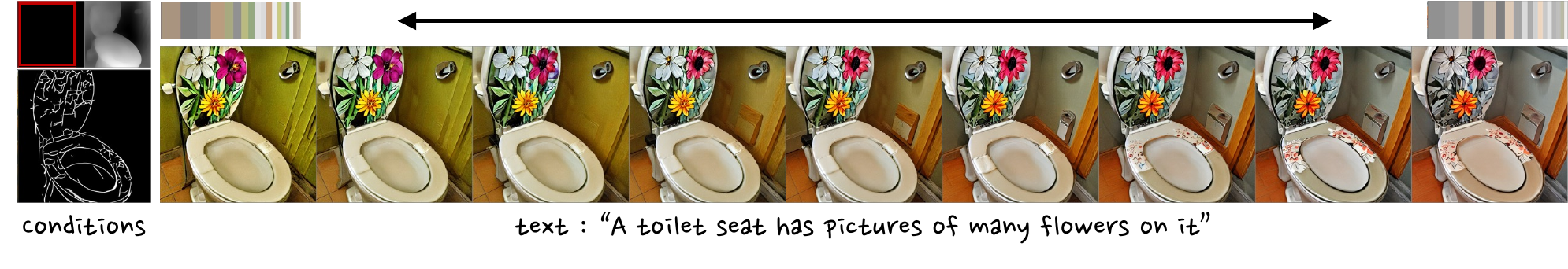}
    \vspace{-2em}
    \caption{\textbf{Interpolation of the attribute modality.} Condition: color palette}
    \label{fig:appx_color_latent_interpolation}
    \vspace{-1.em}
\end{figure*}

\noindent\textbf{Training strategy.}
In Table~\ref{tab:training_ablation}, we analyze the impact of different training strategies. In Config A, all modality types are trained simultaneously. While joint training shows decent performance, it results in a lower Layout score compared to B and C. We hypothesize that the continuous inclusion of the structure modality, which provides rich spatial information, interferes with the effective learning of the layout modality.
To mitigate this, we introduce the structure modality only during the final stage (last 25\%). Interestingly, training structure alone in the final stage (C) achieves better fidelity than training all modalities together (B). This improvement likely arises because the structure modality provides valuable spatial cues for synthesizing high-quality scenes.
In our preliminary tests, we observed similar behavior with ControlNet, where strong structure modalities enhanced the overall fidelity. However, training all modalities together during the late stage risks overlooking structure information, which may lead to slight fidelity degradation.

\noindent\textbf{Inference conditions.}
In Table~\ref{tab:additional_quantitative}, we analyze the performance by varying the inference input conditions. As shown, injecting conditional information related to specific modality types improves the corresponding modality scores. For instance, the attribute modality influences overall image fidelity, as it provides cues for global information.
Similarly, providing conditions such as sketches or depth maps improves the structure and depth scores, respectively. Using all conditions corresponding to structure, layout, and attribute modalities results in significant improvements in both control scores and fidelity (FID). Furthermore, employing mode-specific guidance (MSG) enhances control scores but slightly reduces fidelity. This drop occurs because MSG adjusts (mostly decreases) the influence of attribute and structure, which have a strong impact on fidelity.
Nonetheless, prioritizing user intentions through more precise control makes \ourmodelsc\ a more user-friendly T2I model. For this reason, we use MSG as the default setting in our framework.

\noindent\textbf{Mode-specific guidance (MSG).}
When conditioning on a combination of diverse input modalities, it can be challenging for the model to treat all modalities equally. The proposed MSG alleviates this issue by enabling decoupled guidance control.
In Figure~\ref{fig:modality_guidance} and \ref{fig:appx_modality_guidance}, we present examples of MSG applied to the reference style and depth map conditions. Positive guidance from the reference image adds flowers to the grass, while negative guidance removes them. Similarly, increasing the depth map guidance enhances the smoke effect from a volcano.
Importantly, while MSG improves controllability, information from other modalities, such as boxes or color, remains preserved.

\begin{figure*}[!t]
    \centering
    \includegraphics[width=\linewidth]{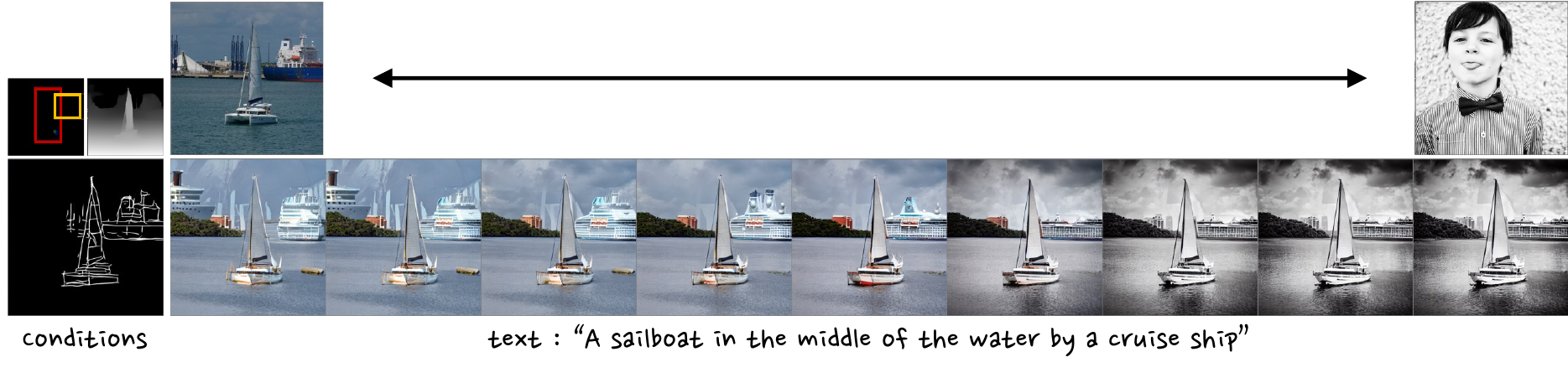}
    \includegraphics[width=\linewidth]{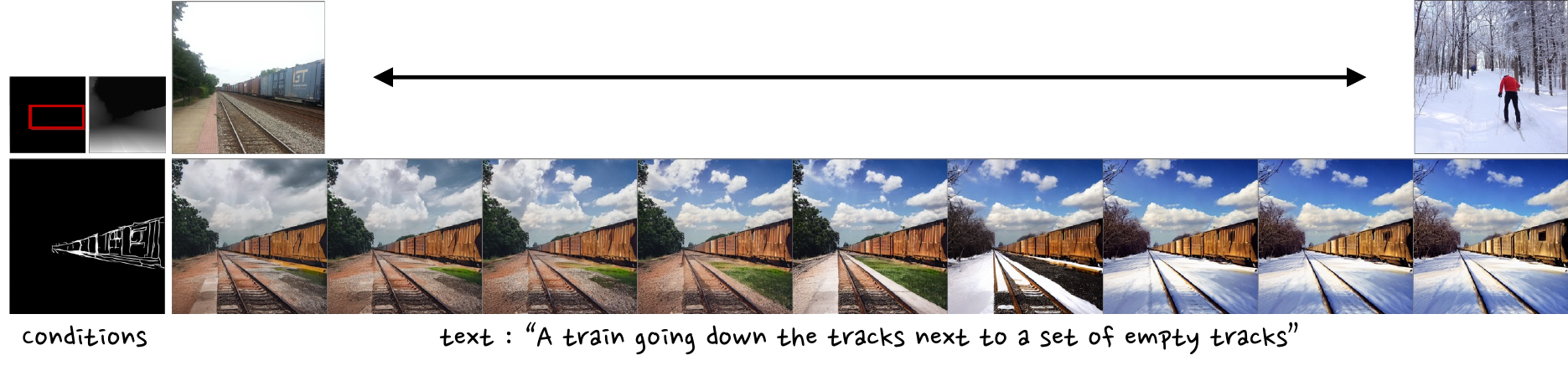}
    \includegraphics[width=\linewidth]{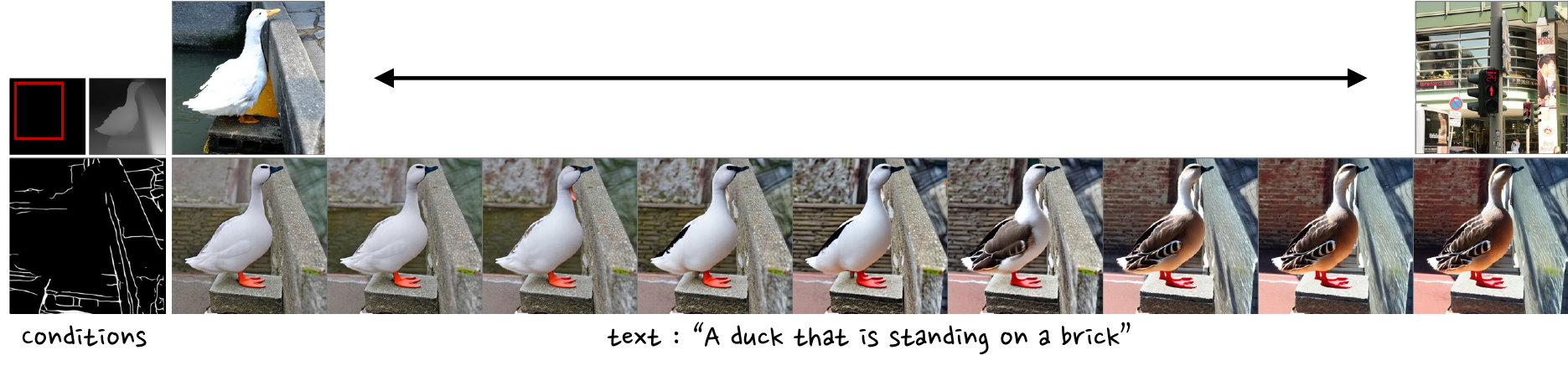}
    \includegraphics[width=\linewidth]{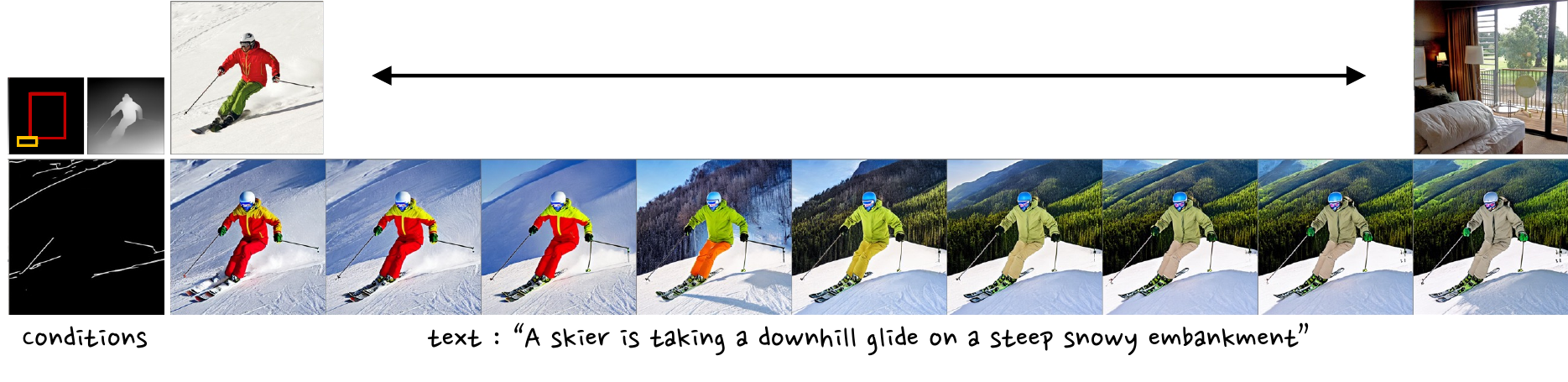}
    \vspace{-2em}
    \caption{\textbf{Interpolation of the attribute modality.} Condition: style image embedding}
    \label{fig:appx_style_latent_interpolation}
    \vspace{-1em}
\end{figure*}

By adjusting the condition through MSG, images are synthesized to reflect the specified condition more or less; however, excessive manipulation can result in a degradation in image quality.
Hence, we shift the mode-specific guidance scale within the range of $[-3.0, 3.0]$ to preserve the image quality, while using a fixed value of 5.0 for the classifier-free guidance scale. Given the sensitivity of the image to sketch guidance, the guidance scale value for this modality is adjusted within the range of $[-1.5, 1.5]$.

\begin{figure*}[t]
    \centering
    \includegraphics[width=\linewidth]{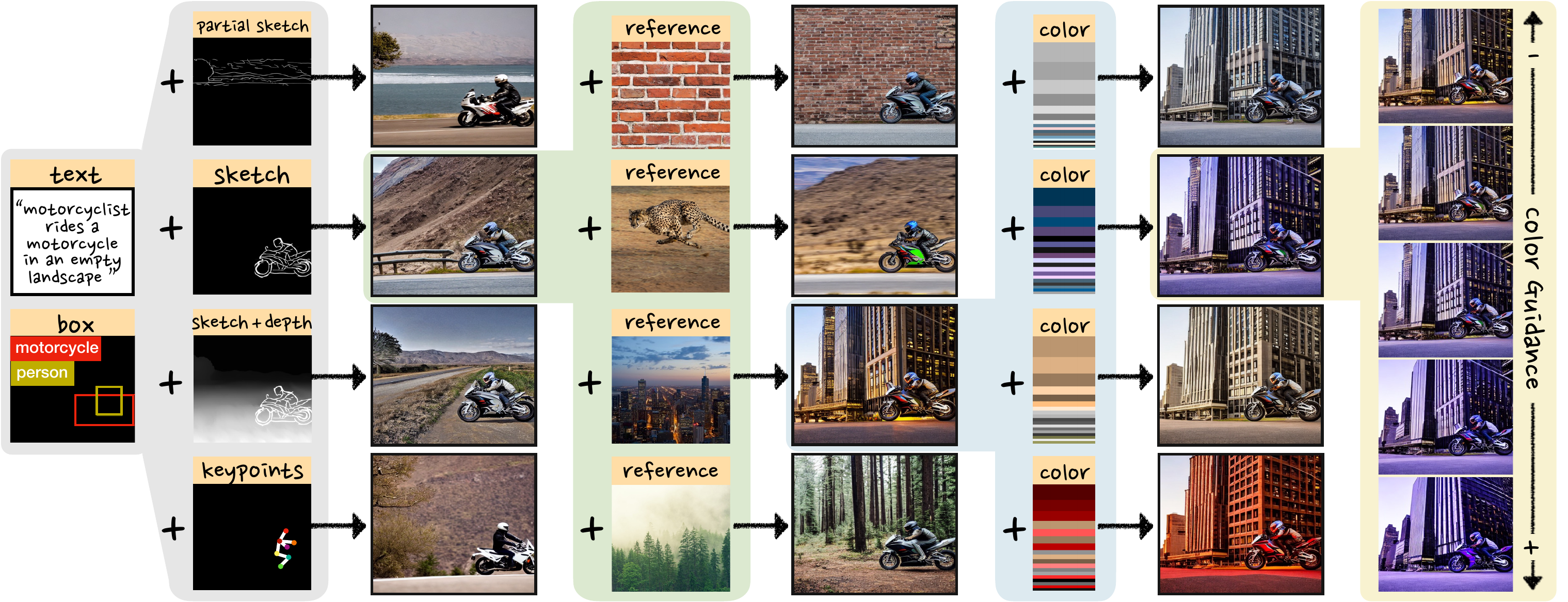}
    
    \caption{\textbf{Versatile applications of {\ourmodelsc}}. {\ourmodelsc} enables flexible manipulation of conditions, providing the image generation aligned with user preferences.  
    Note that all results are generated by our single model at once, not in a sequence.
    }
    
    \label{fig:concept_figure}
\end{figure*}

\begin{figure*}[t]
    \centering
    \includegraphics[width=\linewidth]{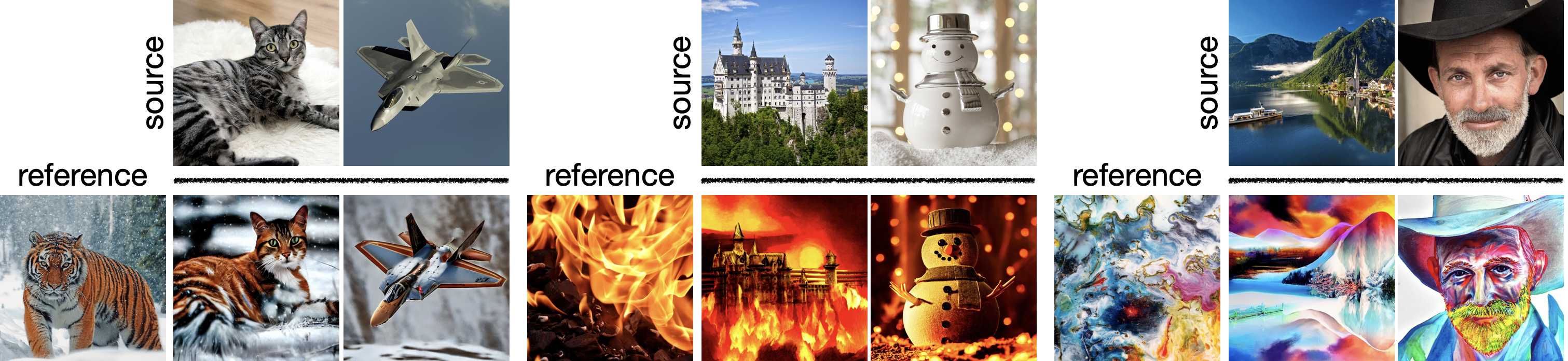}
    \caption{\textbf{Reference-guided and semantic-preserved generation.} 
    Provided with a source image, {\ourmodelsc} demonstrates the results preserving the structure inherent in that image, while effectively incorporating the style elements from a reference image.}
    \label{fig:style_transfer}
     \vspace{-1.5em}
\end{figure*}

\begin{figure*}[t]
    \centering
    \includegraphics[width=\linewidth]{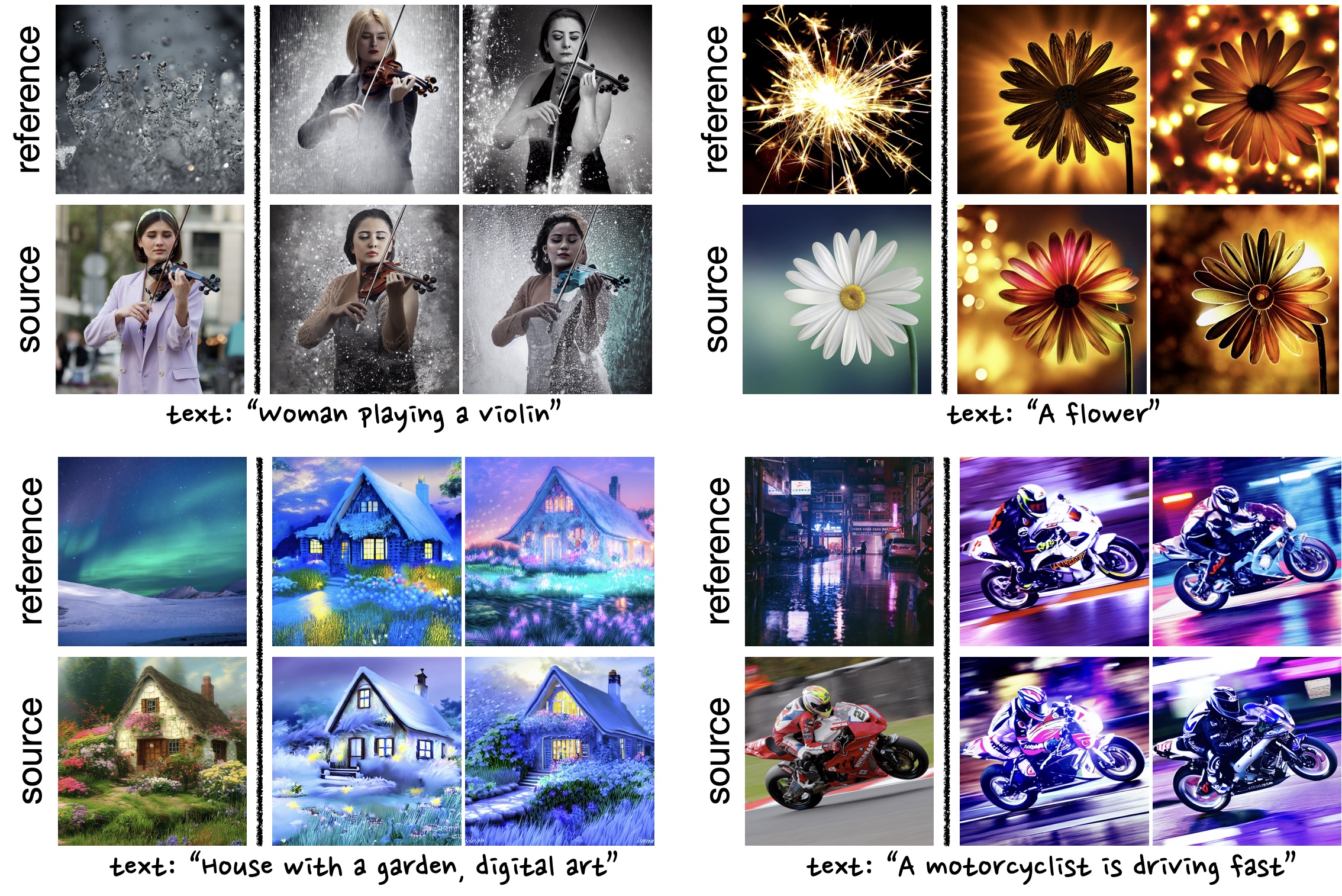}
    \caption{\textbf{Reference-guided and semantic-preserved generation.} {\ourmodelsc} effectively conveys the style from a reference image (top in the first column) while maintaining the structural information of a source image (bottom in the first column).}
    \label{fig:appx_style_transfer}
\end{figure*}

\begin{figure*}[t]
    \centering
    \includegraphics[width=1.0\linewidth]{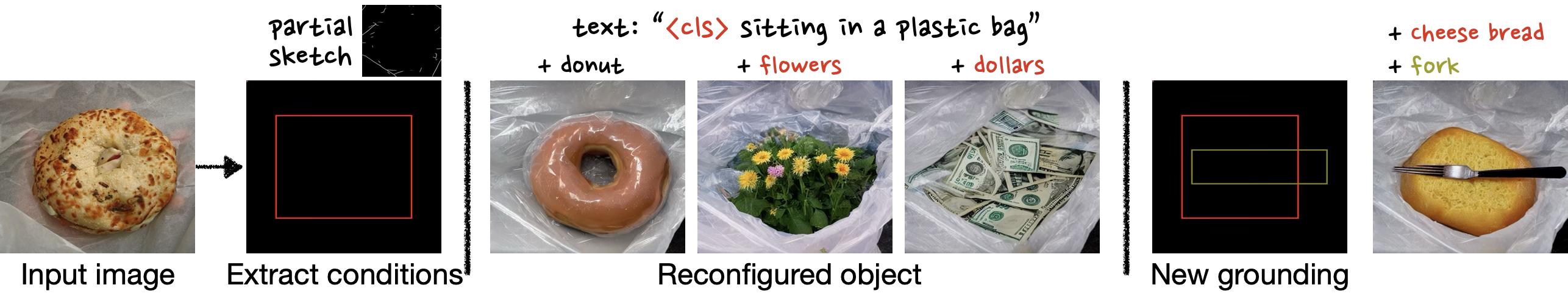}
    
    \caption{\textbf{Object reconfiguration.}
    {\ourmodelsc} reconstructs a new scene by altering the object type or adapting to new grounding information.
    }
    \label{fig:object_reconfiguration}
    \vspace{-1em}
\end{figure*}

Figure~\ref{fig:msg} presents quantitative results demonstrating the effective control achieved with mode-specific guidance (MSG). The results highlight reliable control over image quality and conditioning strength within a reasonable range of the guidance scale.
We observe that layout, structure, and depth scores increase as the guidance scale value rises. Conversely, lowering the scale value reduces the influence of the specified modality without affecting others. Importantly, MSG has minimal impact on FID scores, which remain significantly lower than those of other T2I methods.

\noindent\textbf{Token Selection.}
Our architectural design employs modulated self-attention within the blending blocks. In this setting, we retain only the tokens corresponding to the spatial region of interest after the self-attention operation (e.g. in Equation~\ref{eq:sab}). Table~\ref{tab:token_selection} presents a comparison of three strategies: (1) selecting tokens aligned with the region of interest (our default), (2) using all tokens followed by truncation, and (3) randomly sampling tokens across positions. The results demonstrate that spatially aligned token selection provides better control and structural fidelity.

\noindent\textbf{Structure blending module.}
To evaluate the importance of the structure blending module (SBM), we compare the full DiffBlender model against a variant with the SBM removed. As shown in Table~\ref{tab:sbm}, removing the SBM significantly degrades performance, especially on structure-sensitive metrics such as the Structure and Layout scores. The full model achieves a Layout score of 19.7 and Structure score of 34.8, while the no-SBM variant drops to 15.8 and 31.4, respectively. This highlights the crucial role of SBM in effectively incorporating structural information for accurate spatial and semantic alignment.

\subsection{Applications of \ourmodelsc}

\noindent\textbf{Modality manipulation.}
Unlike structure modality, which require significant human effort to craft conditions, layout and attribute modalities are much easier to manipulate due to their simplicity. In Figure~\ref{fig:latent_interpolation} (a-b), we adjust the layout condition to modify an object’s position or pose. For example, the sun appears in different locations, and the astronaut’s arms are elevated.
In Figure~\ref{fig:latent_interpolation} (c-d), we interpolate between randomly selected color palettes or reference images and apply them to generate new images. Notably, during the smooth transition between color palettes, the seagulls’ positions remain consistent, while the background composition evolves creatively. In the reference interpolation case, substantial transformations occur, particularly in the background, to incorporate the semantics and style of the reference image effectively.
Since \ourmodelsc\ allows users to freely control structure, layout, and attribute modalities when representing a scene, it provides better controllability compared to existing multimodal T2I models that rely on only structure or attribute modalities.

In addition, Figures~\ref{fig:appx_color_latent_interpolation} and \ref{fig:appx_style_latent_interpolation} showcase examples of interpolated results between two conditions from the attribute modality: color palette and reference image.

\noindent\textbf{Multimodal-conditioned generation.} 
Our model demonstrates the ability to generate images by incorporating various types of conditions and can even infer missing modalities when they are not explicitly provided (Figure~\ref{fig:first_teaser}).
The model enables a wide range of application scenarios, as shown in Figure~\ref{fig:concept_figure}. Starting with text and bounding boxes, users can specify the background layout, define detailed human poses, provide reference images for abstract styles, combine colors freely, and adjust the strength of conditions using mode-specific guidance.
With its composable and versatile multimodal design, \ourmodelsc\ allows users to control the synthesized scene flexibly through various modalities in their preferred way.

\noindent\textbf{Reference-guided semantic-preserved generation.}
The preservation of an image's underlying structure while altering its style is a practical challenge~\cite{zhang2023inversion, kwon2022diffusion}.
{\ourmodelsc} adeptly achieves this by extracting the sketch and depth map from a source structure image, as well as the global style and color palette from a reference image.
Combining these conditions, {\ourmodelsc} can synthesize images that blend the semantic and stylistic aspects. In Figure~\ref{fig:style_transfer} and \ref{fig:appx_style_transfer}, {\ourmodelsc} effectively conveys new style onto the source image while maintaining its structural information. Intriguingly, it can seamlessly combine two unrelated images, such as \{jet, tiger\} or \{snowman, fire\} (Figure~\ref{fig:style_transfer}).

\begin{table*}[t]
\caption{
\textbf{Quantitative comparison on COCO17 validation set.} Notably, T2I-Adapter, GLIGEN, and ours are trained on relatively small datasets, whereas ControlNet, Uni-ControlNet, and AnyControl rely on web-scale datasets. Supported modalities are denoted as S (structure), L (layout), and A (attribute). For \# parameters, $^\dagger$ indicates models that support only a single modality per adapter. Enabling all modalities in our benchmark requires training six separate models.
}
\label{tab:quantitative_evaluation}
\vspace{-0.5em}
{
\centering
\setlength{\tabcolsep}{5.1pt}
\begin{tabular}{lllr |ccc |cc}
\hline
Method & Trainset (Size) & \makecell{Supported\\Modality} & \# Params. & Layout $\uparrow$ & Structure $\uparrow$ & Depth $\uparrow$ & FID $\downarrow$ &CLIP $\uparrow$ \\
\hline\hline
ControlNet~\cite{zhang2023@controlnet} & Web (3M) & S & 6$\times$360M$^\dagger$ & 19.6 & 38.0 & 88.8 & 18.0 & 24.9 \\
Uni-ControlNet~\cite{zhao2023unicontolnet} & LAION (10M) & S + A & 880M\phantom{$^\dagger$} & - & \textbf{59.2} & \textbf{91.1} & 20.1 & - \\
AnyControl~\cite{sun2025anycontrol} & MultiGen (3M) & S + A & 790M\phantom{$^\dagger$} & - & - & - & 18.9 & 25.9 \\
\hline
T2I-Adapter~\cite{mou2023@t2iadaptor} & LAION-A (600K) & S & 6$\times$80M$^\dagger$ & 11.2 & 32.1 & 88.2 & 22.4 & \textbf{26.5} \\
GLIGEN~\cite{li2023@gligen} & \textbf{COCO17 (120K)}  & L & 6$\times$210M$^\dagger$ & {17.5} & 30.8 & 87.1 & 19.2 & 24.7 \\
\ourmodelsc & \textbf{COCO17 (120K)} & \textbf{S + L + A} & 450M\phantom{$^\dagger$} & \textbf{20.1} & 37.6 & 88.8 & \textbf{17.0} & 25.3 \\
\hline
\end{tabular}
}
\end{table*}

\begin{figure*}[t]
    \centering
    \includegraphics[width=\linewidth]{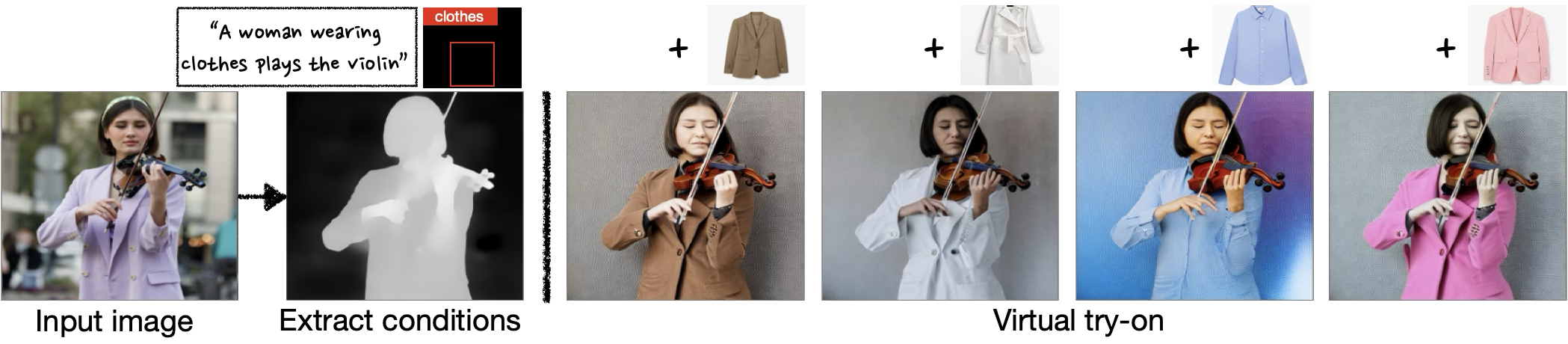}
    
    \caption{\textbf{Virtual try-on application.}
    {\ourmodelsc} can also be applied to virtual try-on scenarios.
}
    \label{fig:vton}
     \vspace{-1em}
\end{figure*}

\noindent\textbf{Object reconfiguration.}
Thanks to the wide range of conditions available in {\ourmodelsc}, it becomes possible to reconstruct scenes containing diverse objects. In Figure~\ref{fig:object_reconfiguration}, we extract a partial sketch of the background and the position of the object from the input image. By altering the prompt for each corresponding object, \eg \textit{flowers} or \textit{dollars}, it can reconfigure the output images by creatively filling in the empty regions. A notable difference between \ourmodelsc's object reconfiguration and traditional inpainting-based image editing~\cite{yang2022@paintbyexample, saharia2022palette, lugmayr2022repaint} lies in the fact that our method can exactly locate the object in the scene, or include new grounding information like a \textit{fork}. 
Moreover, because of its versatile modality adaptation, \ourmodelsc\ can be applied to virtual try-on scenarios (Figure~\ref{fig:vton}).

\subsection{Comparative Study with Previous T2I Methods}
\label{subsec:evaluation}

\noindent\textbf{Baselines.}
We use state-of-the-art conditional T2I models; T2I-Adapter~\cite{mou2023@t2iadaptor}, GLIGEN~\cite{li2023@gligen}, ControlNet~\cite{zhang2023@controlnet}, Uni-ControlNet~\cite{zhao2023unicontolnet}, AnyControl~\cite{sun2025anycontrol}, and PAIR-diffusion~\cite{goel2023@pair}. These models support a range of modalities, from single to multimodal settings.
Note that although Composer~\cite{huang2023@composer} supports both structure and attribute modalities, we exclude it from the comparison due to practical constraints. The authors have not released official code or pretrained models, and the paper lacks quantitative results on standard benchmarks. Moreover, Composer requires training a 5B-parameter model on over 1B image-text pairs, which makes reproduction infeasible due to the substantial computational cost. In contrast, DiffBlender achieves competitive performance with only 450M parameters and a modest training dataset (e.g. COCO).

\begin{table*}[t]
    \centering
    \caption{\textbf{User study questionnaire and results.}
    In total, 42 respondents have participated. 
    In {Question 3}, we provide the user with the originally intended image, extract every conditioning modality from the image, and generate conditional images with possible modalities for each model.
    }
    \vspace{-0.5em}
    \scriptsize
    \label{tab:user_study_main}
    \resizebox{\textwidth}{!}{%
    \begin{tabular}{l|l}
    \hline
    \textbf{Question and and Options} & \textbf{Result} \\
    \hline\hline
    \makecell[l]{\textbf{[Question 1]} Given the color palette, which image most accurately reflects \\ the provided colors? \textbf{[Options]} Generated images of {\ourmodelsc} with or\\ without the color palette condition.} &
    \makecell[l]{with color conditioning (\textbf{81.9\%})\\ without color conditioning (18.1\%)} \\
    \hline
    \makecell[l]{\textbf{[Question 2]} Given the reference image, which image most accurately \\ reflects the style of the reference image? \textbf{[Options]} Generated images from \\ PAIR-diffusion and {\ourmodelsc}.} & 
    \makecell[l]{PAIR-diffusion (11.4\%),\\{\ourmodelsc} (\textbf{88.6\%})} \\
    \hline
    \makecell[l]{\textbf{[Question 3]} Given the user's intended image, which image most accurately\\ reflects the user's intention? \textbf{[Options]} Generated images from SD ver.1.4,\\ ControlNet, GLIGEN, T2I-Adapter, and {\ourmodelsc}.} &
    \makecell[l]{SD ver.1.4\,(0.0\%), ControlNet\,(18.6\%),\\ GLIGEN\,(0.0\%), T2I-Adapter\,(15.7\%),\\ {\ourmodelsc}\,(\textbf{65.7\%})} \\
    \hline
    \makecell[l]{\textbf{[Question 4]} (1) Does mode-specific guidance change the output images?\\(2) Is the change reasonable according to the specified modality? (3) During\\ the guidance, are the other modalities being maintained?} &
    \makecell[l]{(1) Yes (\textbf{96.8\%}), No (3.2\%)\\ (2) Yes (\textbf{95.2\%}), No (4.8\%)\\ (3) Yes (\textbf{97.6\%}), No (2.4\%)} \\
    \hline
    \end{tabular}
    }
\end{table*}

\begin{figure}[t]
    \centering
    \includegraphics[width=\linewidth]{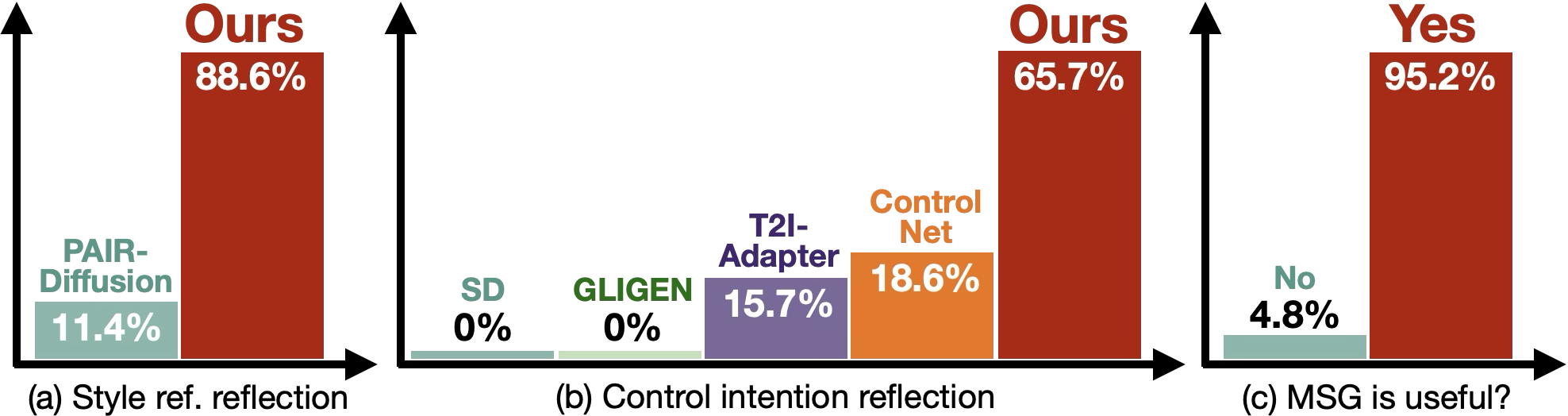}
    
    \caption{\textbf{User study.} }
        \label{fig:user_study}
        \vspace{-1em}
\end{figure}

\begin{figure*}[t]
    \centering
    \includegraphics[width=\linewidth]{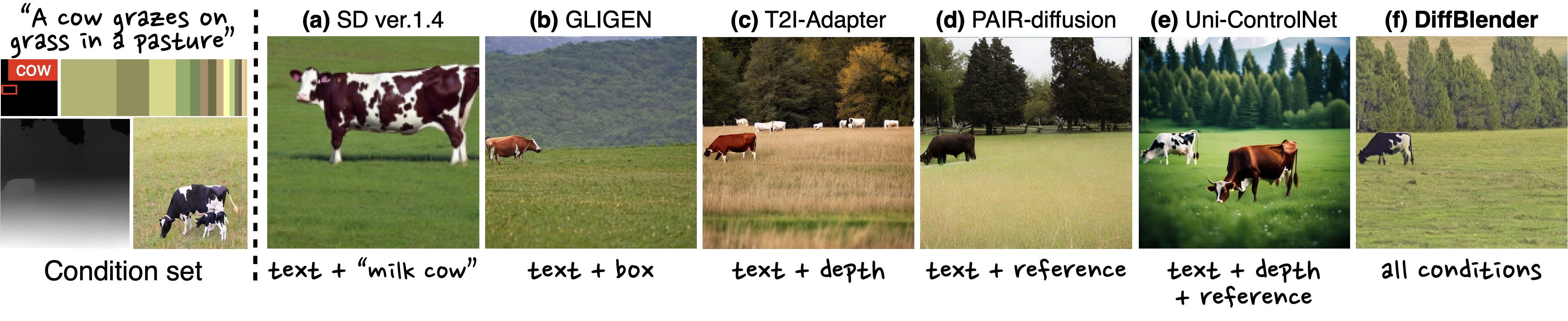}
    
    \caption{\textbf{Visual comparison with previous T2I methods.}
    Providing the conditions on the left, we present SD~\cite{rombach2022@ldm}, GLIGEN~\cite{li2023@gligen}, T2I-Adapter~\cite{mou2023@t2iadaptor}, PAIR-diffusion~\cite{goel2023@pair}, Uni-ControlNet~\cite{zhao2023unicontolnet}, and {\ourmodelsc} results.
          }
    \label{fig:qualitative_comparison}
\end{figure*}

\noindent\textbf{Quantitative evaluation.}
As shown in Table~\ref{tab:quantitative_evaluation}, {\ourmodelsc} achieves not only high fidelity (FID and CLIP scores) but also reliable performance in multi-conditional generation (Layout, Structure, and Depth scores).
Although \ourmodelsc\ does not perform the top scores, it delivers competitive results with a substantially smaller dataset. While Uni-ControlNet and AnyControl demonstrates strong performance, they require 25-100 times more data than ours, which poses significant challenges for modality extension or training in non-natural image domains.
Notably, \ourmodelsc\ is the only model that supports all modality types within a single framework. It processes all modality conditions through a unified 450M-parameter module, whereas ControlNet, T2I-Adapter, and GLIGEN require multiple adapters to support diverse modalities, making them less practical.
Thanks to its effective modular design for supporting all-in-one modalities, \ourmodelsc\ achieves excellent controllability and fidelity, even with limited data.

\noindent\textbf{User study.}
In the user study, \ourmodelsc\ demonstrates strong performance in reflecting user intentions (Figure~\ref{fig:user_study}).
\tab\ref{tab:user_study_main} shows details of questionnaire and answers used in Figure~\ref{fig:user_study}.
\textbf{Questions 1 and 2} focus on evaluating the faithfulness of the attribute modality, such as color palette and reference image embedding. Results show that {\ourmodelsc} accurately captures the color palette (81.9\%) and faithfully reflects the style of the reference image (88.6\%).
\textbf{Question 3} assesses the faithfulness of the layout modality. The results indicate that {\ourmodelsc} excels at customizing image generation according to user intentions (65.7\%), thanks to its ability to condition on diverse modalities.
\textbf{Question 4} investigates the effectiveness of mode-specific guidance (MSG) in reliably controlling the specified modality while maintaining the integrity of others.

For \textbf{Questions 1–3}, five examples and corresponding answers were recorded for each question, while three examples and answers were recorded for \textbf{Question 4}. No personally sensitive or vulnerable data were collected during the study.
We excluded Uni-ControlNet~\cite{zhao2023unicontolnet} from our comparison due to its primary emphasis on the structure modality. While it includes a global condition (content), its FID score is significantly higher (24.0) when the content condition is used, compared to {\ourmodelsc}, which achieves a superior FID score of 17.5 with a reference condition. Additionally, Uni-ControlNet’s local conditioning module for image-form inputs (structure modality) closely resembles ControlNet models. For these reasons, Uni-ControlNet was not included in the human evaluation.

\noindent\textbf{Qualitative evaluation.}
As shown in Figure~\ref{fig:qualitative_comparison}, we provide a visual comparison, where the samples are generated using the available condition(s) for each model.
SD exhibits the ability to generate high-quality result but lacks precision in accurately guiding the object and background layout.
Single-modal models (GLIGEN and T2I-Adapter) are better but they have limitation in capturing the fine details (\eg structure).
PAIR-diffusion can condition on the reference image, but its usage is bounded because it requires input pair images sharing identical semantic objects.
Uni-ControlNet, while reflecting multiple modalities simultaneously, often generates overly saturated and unrealistic images due to the unbalanced modality influence (mostly attribute).
In contrast, {\ourmodelsc} has the advantage of utilizing various conditions within a single model, and we can specify not only the layout but also detailed poses or reference image information. 

\begin{figure}[t]
    \centering
    \includegraphics[width=\linewidth]{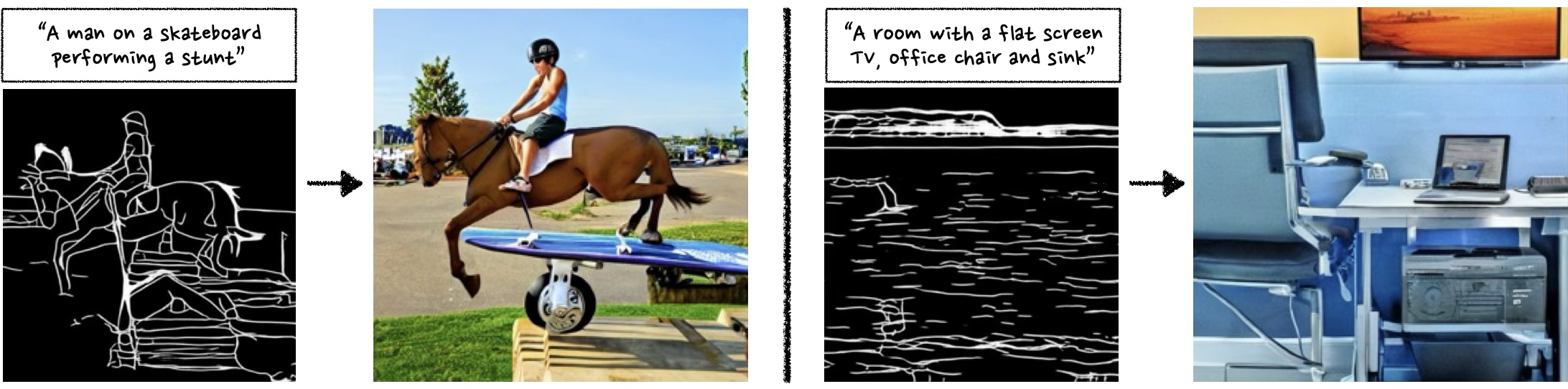}
    
    \caption{\textbf{Failure cases of {\ourmodelsc}.} }
        \label{fig:failure}
\end{figure}

\section{Conclusion}
\label{sec:conclusion}

We present \ourmodelsc, a novel diffusion-based multimodal text-to-image (T2I) generation model capable of handling diverse conditioning modalities. To enable this, we categorize and redefine commonly used modalities into three types—structure, layout, and attribute—and develop corresponding conditioning modules that support efficient and composable training. \ourmodelsc\ achieves high-quality image synthesis by faithfully integrating complex and heterogeneous conditions, making it well-suited for a broad range of practical applications.

\noindent\textbf{Limitations.}
As \ourmodelsc\ is designed to process multimodal inputs, it may encounter difficulties when the provided conditions are unaligned. As illustrated in Figure~\ref{fig:failure}, the model attempts to preserve the input sketch while simultaneously introducing objects specified in the text prompt. This mismatch between modalities can degrade image quality, as the model struggles to reconcile conflicting cues into a coherent and visually consistent output.

{
\small
\bibliographystyle{ieeenat_fullname}
\bibliography{reference}
}

\end{document}